\documentclass[journal]{IEEEtran}
\usepackage{color}
\usepackage{graphicx}
\graphicspath{{./Pics/}}
\DeclareGraphicsExtensions{.pdf,.jpeg,.png}
\usepackage{multirow}
\usepackage{tabularx}
\usepackage{booktabs}
\usepackage{amsmath}
\usepackage{amssymb}
\usepackage[ruled,vlined]{algorithm2e}
\usepackage{algorithmic}
\ifCLASSINFOpdf
\else
\fi

\makeatletter
\def\endthebibliography{%
  \def\@noitemerr{\@latex@warning{Empty `thebibliography' environment}}%
  \endlist
}
\makeatother
\begin{document}
%
\title{Self-Enhanced Convolutional Network for\\ 
Facial Video Hallucination}
%
%
%

\author{
        Chaowei~Fang,
        Guanbin~Li, \emph{Member, IEEE},
        Xiaoguang~Han, \emph{Member, IEEE},
        and~Yizhou~Yu, \emph{Fellow, IEEE}
\thanks{This work was partially done by C. Fang when he was a PhD candidate at the Department of Computer Science, The University of Hong Kong, Hong Kong.}
\thanks{G. Li is with the School of Data and Computer Science, Sun Yat-sen University, Guangzhou 510006, China.}
\thanks{X. Han is with the Shenzhen Research Institute of Big Data, The Chinese University of Hong Kong (Shenzhen), Shenzhen 518172, China.}
\thanks{Y. Yu is with the Department of Computer Science, The University of Hong Kong, Hong Kong (Email: yizhouy@acm.org). }
\thanks{This work was partially supported Hong Kong Research Grants Council under Research Impact Fund (R-5001-18), National Natural Science Foundation of China under Grant No.61976250 and No.U1811463, Fundamental Research Funds for the Central Universities under Grant No.18lgpy63, grants No.2018YFB1800800, No.2018B030338001, NSFC-61629101, No.ZDSYS201707251409055 and No.2017ZT07X152 (Corresponding author: Yizhou Yu). }
}


\markboth{IEEE Transactions on Image Processing,~Vol.~xx, No.~x, xxxx~20xx}%
{Shell \MakeLowercase{\textit{Fang et al.}}: Self-Enhanced Convolutional Network}

\maketitle

\begin{abstract}
 As a domain-specific super-resolution problem, facial image hallucination has enjoyed a series of breakthroughs thanks to the advances of deep convolutional neural networks. However, the direct migration of existing methods to video is still difficult to achieve good performance due to its lack of alignment and consistency modelling in temporal domain.
 Taking advantage of high inter-frame dependency in videos, we propose a self-enhanced convolutional network for facial video hallucination. It is implemented by making full usage of preceding super-resolved frames and a temporal window of adjacent low-resolution frames. Specifically, the algorithm first obtains the initial high-resolution inference of each frame by taking into consideration a sequence of consecutive low-resolution inputs through temporal consistency modelling. It further recurrently exploits the reconstructed results and intermediate features of a sequence of preceding frames to improve the initial super-resolution of the current frame by modelling the coherence of structural facial features across frames.
 Quantitative and qualitative evaluations demonstrate the superiority of the proposed algorithm against state-of-the-art methods. Moreover, our algorithm also achieves excellent performance in the task of general video super-resolution in a single-shot setting.
\end{abstract}


\IEEEpeerreviewmaketitle

 \IEEEPARstart{F}{ace} hallucination, also known as face super-resolution (SR), is a fundamental problem in computer vision because of its vast application scenarios, such as video surveillance, facial attribute analysis and visual content enhancement. Recently reconstructing static high-resolution (HR) face images from low-resolution (LR) ones has been widely studied ~\cite{shi2019face,bulat2018super}. 
 However the development of SR techniques in facial videos is far less explored due to its high complexity and requirement in effective spatio-temporal modelling. In this paper we focus on hallucinating high-resolution (HR) videos of talking faces from low-resolution (LR) ones. Faces in most of such videos do not have large or sudden motions, but with relatively small rotations.

\begin{figure}[t]
\centering
\includegraphics[width=\columnwidth]{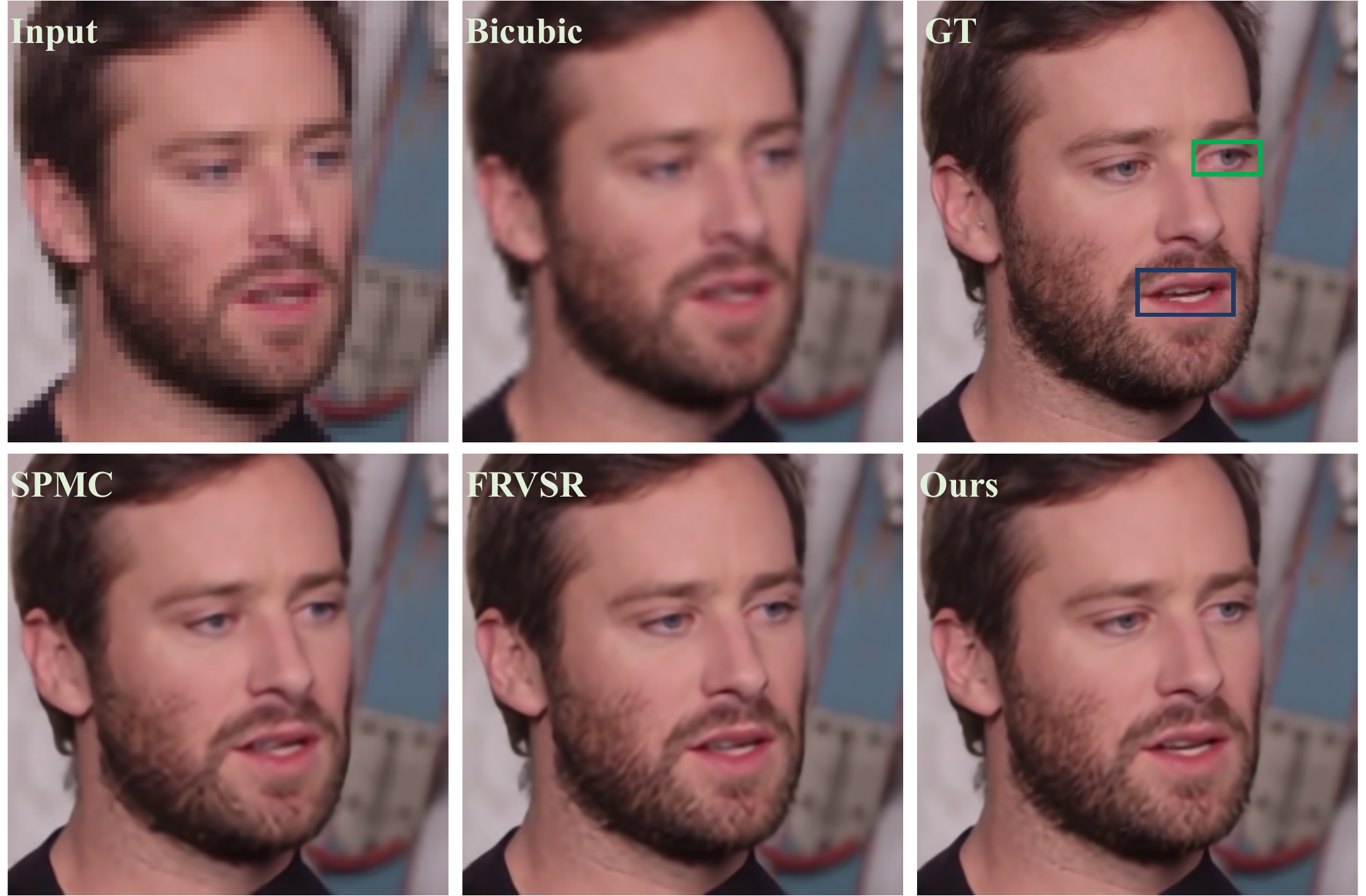}
\caption{Our method aims to generate high-resolution facial video frames from low-resolution inputs. Taking a specific frame as an example, existing video super-resolution methods SPMC~\cite{tao2017detail} and FRVSR~\cite{sajjadi2018frame} have difficulty in recovering components with complicate structures such as the right eye (green box) and teeth (blue box). Our method is capable of achieving more promising results. The input image is visualized using pixel duplication. `GT' represents ground truth image. (Best viewed in close-up) }
\label{fig:topic}
\end{figure}

Video super-resolution is a notorious ill-posed problem. The challenge of this problem resides in restoring individual frames with high-definition appearances while requiring natural inter-frame consistency and visual friendliness. Additionally a video SR method should exploit effective and relevant information from the rest of the video for signal reconstruction. Traditional methods~\cite{liu2014bayesian,ma2015handling} mainly focus on reconstructing HR images via estimating blur kernel, inter-frame flow fields and extra noises. With the development and wide application of deep learning techniques, CNN-based methods turn out to be the mainstream in video super-resolution, which are significantly superior to traditional methods. One of the most intuitive solutions is to perform the restoration of the current frame by registering other adjacent frames and using CNN based feature fusion~\cite{kappeler2016video,caballero2017real,tao2017detail}. Unfortunately, this kind of method has the following shortcomings: 1) it is usually arduous to accurately register two frames within a long time interval which is highly likely to have a negative impact on subsequent fusion; 2) the fusion of all frames leads to a sharp increase in the amount of computation, which greatly reduces the overall efficiency; 3) using relatively small number of frames ignores much spatial and temporal information which could be otherwise very helpful. When directly applied to facial video hallucination, existing state-of-the-art video SR methods~\cite{tao2017detail,sajjadi2018frame} can successfully generate temporally coherent results with acceptable appearances in smooth regions such as cheek and nose. However they are not competent for the super-resolution reconstruction of image components with relatively complicate structures or textures, for example the regions of eyes and teeth as shown in Fig.~\ref{fig:topic}.

To address the above issue, we present a so-called~\emph{Self-Enhanced Convolutional Network}, which is a novel end-to-end learning framework and can fully exploit both long-term spatial and temporal information for enhancing the hallucination inference of later frames. The self-enhancement of our method is inspired by the following two perspectives.  1) The spatial information of the preamble frames are crucial for restoring subsequent frames as there is a large amount of inter-frame redundancy especially in facial videos. Thus multiple super-resolved results of previous frames are propagated to enhance the prediction of subsequent frames. 2) Considering temporal information is paramount to reason the appearance of later frames, ConvLSTM~\cite{xingjian2015convolutional} is applied to enhance the feature representation of every frame by sequentially encoding the features from the past frames. The self-enhancement model is implemented through an encoder-decoder architecture. It absorbs in an initial prediction of current frame and registered HR frames of past frames, resulting in a refinement map of the initial prediction. The feature representation of each frame resides in-between the encoder and decoder. To involve in information of future frames and further boost the SR performance, neighboring LR images are used to generate the initial HR estimation for each frame via a local frame fusion network.  Except for facial videos, our proposed method also has a strong advantage for the super-resolution of general videos as it is particularly good at learning and capturing structural and temporal consistency, especially for the reconstruction of scenes with intricate and trifling structures (e.g. buildings).

In summary, this paper has the following contributions.
\begin{itemize}
\item A self-enhanced convolutional network is proposed for facial video hallucination. The uniqueness of our model is that it makes use of both spatial and temporal information across all preceding frames.
\item Three ConvLSTM-based recurrence strategies are devised to excavate temporal information for enhancing the feature representation of every frame. 
\item Our proposed method has achieved state-of-the-art performance in: two facial video datasets, VoxCeleb~\cite{Nagrani17} and RAVDESS~\cite{livingstone2012ravdess}; two single-shot generic video datasets, VID4~\cite{liu2014bayesian} and Harmonic collected from the Internet.
\end{itemize}

\section{Related Work}
Image/video super resolution has been studied for a long time. We refer to~\cite{schulter2015fast} for a detailed survey. In this section, we mainly discuss the related works based on deep learning.

\subsection{Image/video Super-Resolution}
The basic idea of recent deep learning based methods~\cite{dong2016image,ledig2017photo} is to design a CNN architecture to map low resolution images to their HR versions. In \cite{pathak2018efficient}, an improved version of \cite{ledig2017photo} is proposed with the help of convolutional non-local operation~\cite{wang2018non}. A novel super-resolution method is developed in \cite{wang2018esrgan} using residual-in-residual dense blocks~\cite{huang2017densely}. During the training stage, Relativistic GAN~\cite{jolicoeur2018relativistic} is employed for achieving realistic predictions. Video super resolution, as an extension of image super resolution, attracts more attentions for its practicality but being more challenging. To extend single frame SR model to its multi-frame version, \cite{kappeler2016video} attempts to utilize multiple motion compensated frames/features when super-resolving each frame. \cite{caballero2017real} and \cite{tao2017detail} utilize consecutive neighboring frames to produce the super-resolution output of the current frame with the help of flow based motion correction. A joint upsampling and warping operation~\cite{makansi2017end} is proposed for fusing neighboring frames in video super-resolution. \cite{jo2018deep} super-resolves every LR image using multiple frames via learning a dynamic upsampling filter for each pixel in the target HR image and a residual image. \cite{Liu_2017_ICCV} devises a multi-scale temporal adaptive neural network and a spatial alignment network for utilizing inter-frame dependency in video super-resolution. Considering high inter-frame repeatability in videos, a frame recurrence strategy is proposed in~\cite{sajjadi2018frame} to propagate the spatial information of previously estimated HR frames to all subsequent frames and enhance their HR predictions. As only one previous frame is fused into the inference procedure of current frame, temporal connection across frames is weak which might miss lots of inter-frame spatial dependencies, especially information provided by future frames. Our method differs it from two perspectives. First, multiple neighboring LR frames are utilized to generate an initial prediction through a local frame fusion network. Second, ConvLSTM-based recurrence module is devised to enhance the feature representation of every frame. \cite{huang2015bidirectional} devises a bidirectional recurrent convolutional network to learn long-term temporal and contextual information for video super-resolution. But each frame relies on intermediate features from both past and subsequent frames. All images in the input clip are required to be processed simultaneously and the memory cost grows linearly with respect to its length. Our devised bidirectional recurrent module avoids this shortcoming as each input frame can be super-resolved independently after obtaining features of previous frames.
\begin{figure*}[t]
\centering
\includegraphics[width=\linewidth]{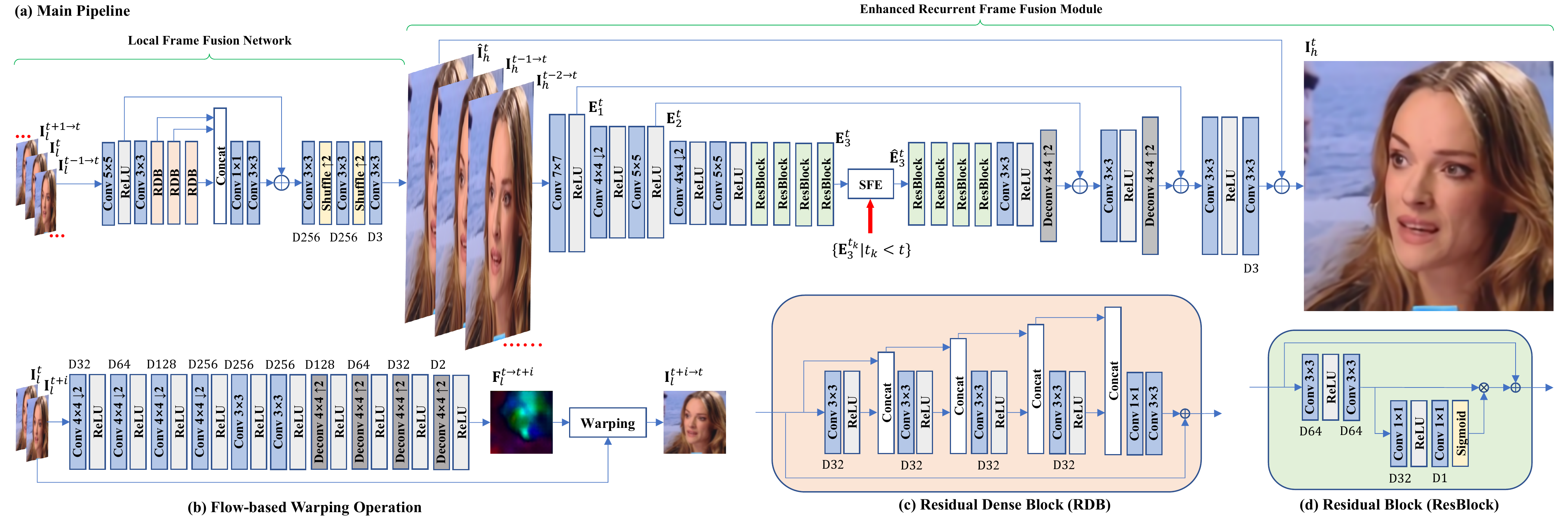}
\caption{Our network architecture for super-resolving $64 \times 64$ images to $256 \times 256$ ones. `D[$D$]' represents a convolution or deconvolution layer with the number of output channels set to $D$. The $x\times y$ beside the box of convolution/deconvolution layer indicates kernel size. `$\uparrow 2$' means the deconvolution layer upsample the feature map to 4 times while `$\downarrow 2$' means the convolution layer downsample the feature map to one quarter. The pixel shuffle layer rearranges the input $s^2D\times H \times W$ tensor to a $D\times sH \times sW$ tensor. $s=2$ for each shuffle layer. The main pipeline (a) of our method consists of two stages. The first one utilizes a local frame fusion network  built on residual dense blocks (c) to produce an initial super-resolution inference. The second stage takes advantage of the previously estimated results and encodes feature maps of past frames sequentially (SFE) to enhance the feature representation of current frame. The adopted residual block is shown in (d).} The optical flow module is shown in (b).
\label{fig:model}
\end{figure*}

\subsection{Face hallucination}
As a special case of image/video super-resolution, facial image hallucination has drawn much more attentions due to its wider application scenarios. Most deep neural network based methods attempt to integrate facial prior knowledge into the CNN architectures. \cite{zhou2015learning,tuzel2016global} implicitly exploit global facial features learned using fully connected layers. \cite{shi2019face} utilizes a reinforcement learning policy to generate HR face image patch by patch iteratively. \cite{chen2018fsrnet,bulat2018super,song2017learning} explicitly make use of facial priors (landmarks/parsing maps) to help inferring the restoration of HR face images or training neural networks. Wavelet coefficients of HR images are inferred from the embedded features of the low resolution faces and are then used to reconstruct the expected HR image~\cite{huang2017wavelet}. Based on PixelCNN~\cite{oord2016pixel}, a novel face image super-resolution model~\cite{dahl2017pixel} is set up to recurrently reconstruct every pixel. This method is hard to restore images with large spatial sizes because of its computational cost.  On the other hand, generative adversarial models~\cite{goodfellow2014generative} are widely used in face super-resolution. UR-DGN~\cite{yu2016ultra} is claimed to be the first face SR method using generative adversarial network. \cite{chen2017face} discusses the efficacy of Wasserstein GAN~\cite{arjovsky2017wasserstein,gulrajani2017improved} in training face SR networks. \cite{xu2017learning} learns a CNN model to super-resolve blurry face and text images with a complicated objective function consisting of pixel-wise MSE, feature matching and adversarial loss. As far as we know, no literature published by conference/journal on face video SR using deep learning is found. Without additional constraints, single face image SR methods can hardly work well in face video SR because of the deformity to guarantee smoothness across frames.
\section{Method}
Denote a sequence of facial video frames as $\{\mathbf X^t\}$, where $t\in[1,N]$ and $t\in\mathcal{N}_+$. $\mathbf X^t$ is a single frame with resolution $w\times h$. Facial video hallucination aims to generate the high-resolution counterpart $\mathbb Y=\{\mathbf Y^t\}$ composed of frames with resolution $rw\times rh$ where $r$ is the upscaling factor. In the following, we first give an overview of our proposed network architecture and then describe the details of each module.

\subsection{Self-Enhanced Convolutional Architecture}\label{sec:overview}
The overall architecture of our proposed self-enhanced convolutional network is illustrated in Fig. \ref{fig:model}. The base of our method consists of two cascaded subnetworks. The first subnetwork is named as local frame fusion network, which takes multiple aligned neighbouring LR frames as input and aims at generating an initial super-resolved result for each independent frame. The second subnetwork is named as enhanced recurrent frame fusion module which refines the result of the local frame fusion network with the help of aligned super-resolved images and features from previous frames.

\subsection{Local Frame Fusion Network}\label{sec:lffn}
For sake of involving in information of future frames and providing a high starting point for subsequent subnetwork, we set up a local frame fusion network based on the residual dense network (RDN~\cite{zhang2018residual}).  $2T_1+1$ consecutive LR frames $\{\mathbf X^{k} |k\in[t-T_1,t+T_1]\}$ are used as the input when super-resolving frame $t$. First of all, to make up inter-frame differences caused by facial/camera motions, an optical flow module is exploited to warp every LR image $\mathbf X^{k}$ to frame $t$ as shown in Fig. \ref{fig:model} (b). Practically the optical flow field $\mathbf F^{t\rightarrow k}$ from $\mathbf X^{t}$ to $\mathbf X^{k}$ is used to bi-linearly sample an aligned counterpart of $\mathbf X^{k}$. We define the warped result of $\mathbf X^{k}$ as $\mathbf X^{k\rightarrow t}$. The mean square error loss with total variation regularization is imposed on the optical flow module,
\begin{equation}
L_f^{t,k} = \frac{1}{cwh}\|\mathbf X^{k\rightarrow t}-\mathbf X^t\|_{2}^2+\frac{\alpha}{2wh}(\|\nabla_x \mathbf F^{t\rightarrow k}\|_2^2+\|\nabla_y \mathbf F^{t\rightarrow k}\|_2^2),
\end{equation}
\noindent where $\alpha$ is a constant and $c$ is the channel of input image. $\nabla_x$ and $\nabla_y$ are horizontal and vertical derivation operation respectively. The overall loss function for training the optical flow network is as follows,
\begin{equation} \label{eq:loss-flow}
L_f^t=\frac{\gamma}{2T_1}\sum_{k=t-T_1,k\neq 0}^{t+T_1} L_f^{t,k}.
\end{equation}

RDN is a state-of-the-art SR method for static image SR method. We extend it into a multi-frame version by replacing the single input image with the concatenation of aligned LR images $\{\mathbf X^{k\rightarrow t}\}$.
Let the output be $\mathbf{\hat Y}^t$. We use the following loss function for training the local frame fusion network,
\begin{equation}\label{eq:loss-lffn}
L_l^{t} = \frac{1}{cr^2wh}\|\mathbf{\hat Y}^t - \mathbf G^t\|_2^2,
\end{equation}
\noindent where $\mathbf G^t$ represents the ground-truth image of frame $t$. Details about the network architecture and residual dense blocks are presented in Fig. \ref{fig:model} (a) and (c).

\subsection{Enhanced Recurrent Frame Fusion Module}\label{sec:erffn}
\begin{figure*}[t]
\centering
\includegraphics[width=\linewidth]{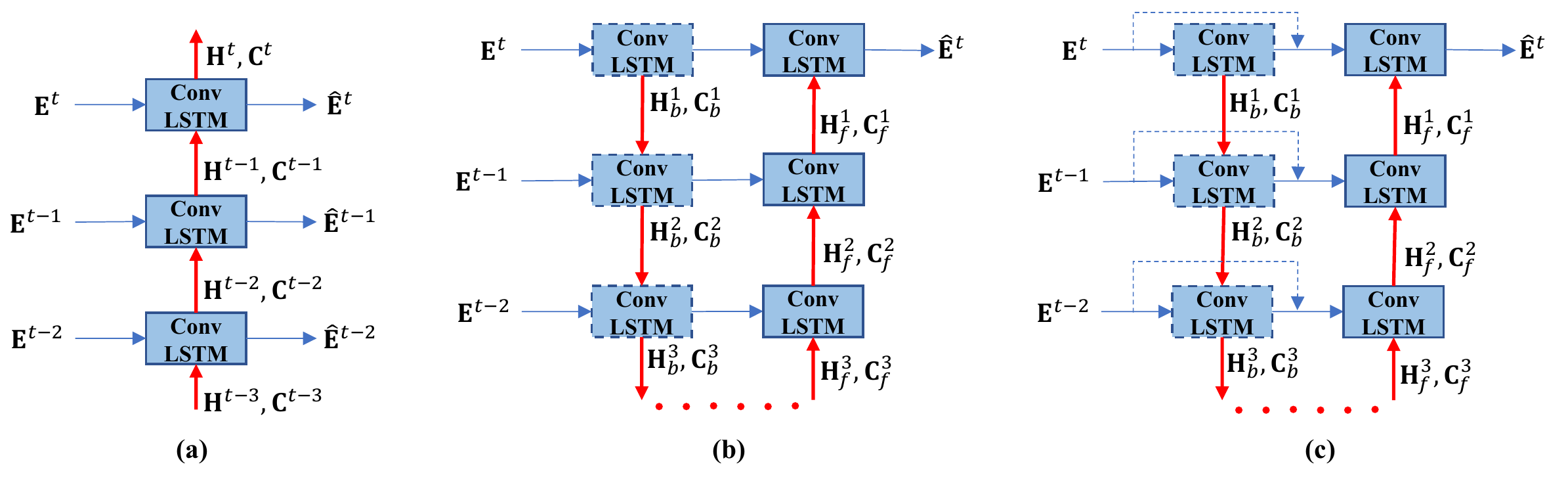}
\caption{Three sequential feature encoding strategies. The left (a) propagates the cell output and hidden variable to next frame recurrently. The second (b) collects $T_3$ features from past frames and employs bidirectional ConvLSTM to encode them at every frame. In the third (c) strategy, the input feature maps are also fed into the forward pass.}
\label{fig:sqe}
\end{figure*}

\begin{figure}[t]
\centering
\begin{minipage}{0.9\columnwidth}
\centering
\includegraphics[width=0.32\textwidth]{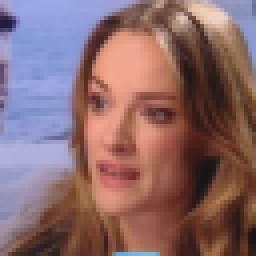}
\includegraphics[width=0.32\textwidth]{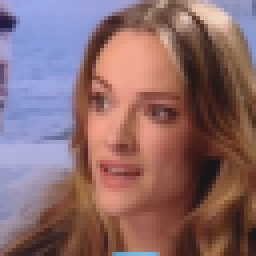}
\includegraphics[width=0.32\textwidth]{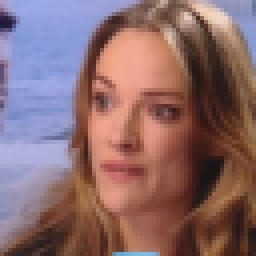}
\end{minipage}\vspace{1mm}
\begin{minipage}{0.9\columnwidth}
\centering
\includegraphics[width=0.32\textwidth]{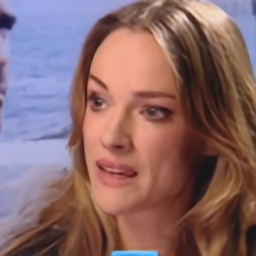}
\includegraphics[width=0.32\textwidth]{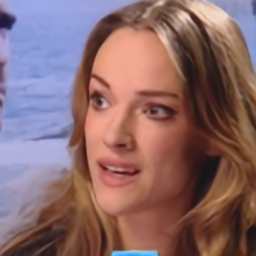}
\includegraphics[width=0.32\textwidth]{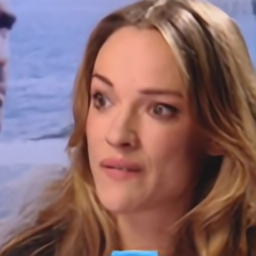}
\end{minipage}\vspace{1mm}
\begin{minipage}{0.9\columnwidth}
\centering
\includegraphics[width=0.32\textwidth]{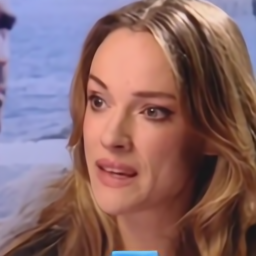}
\includegraphics[width=0.32\textwidth]{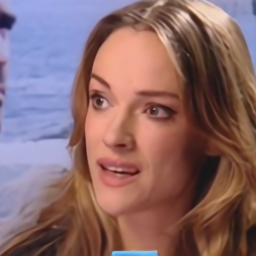}
\includegraphics[width=0.32\textwidth]{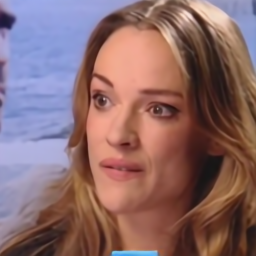}
\end{minipage}
\caption{Given a sequence of low-resolution images (1st row), preliminary HR images are generated using the local frame fusion network (2nd row). Then it is refined by our enhanced recurrent frame fusion module (3rd row).}
\label{fig:stages}
\end{figure}

\begin{algorithm}[t]
\caption{One-way ConvLSTM.}
\fontsize{8pt}{10pt}
\selectfont
\KwIn{ Feature of current frame: $\mathbf{E}^{t}$; \\
 ConvLSTM variables of previous frame: $\mathbf{H}^{t-1}$, $\mathbf{C}^{t-1}$; \\
 ConvLSTM parameters: $\theta_l$.
}
\KwOut{Enhanced feature: $\mathbf{\hat E}^{t}$.}
\begin{algorithmic}[1]
\STATE $\mathbf{H}^{t},\mathbf{C}^{t}  =  \mbox{ConvLSTM}(\mathbf{H}^{t-1},\mathbf{C}^{t-1}, \mathbf{E}^{t}, \theta_l)$.
\STATE $\mathbf{\hat E}^{t} \gets \mathbf H^t$.
\end{algorithmic}
\label{alg:convlstm-one}
\end{algorithm}

\begin{algorithm}[t]
\caption{Cascaded Bidirectional ConvLSTM.}
\fontsize{8pt}{10pt}
\selectfont
\KwIn{ Features: $\{\mathbf{E}^{k}|k=t,\cdots, t-T_t+1\}$; \\
 ConvLSTM parameters: $\theta_l^b$, $\theta_l^f$. }
\KwOut{Enhanced feature: $\mathbf{\hat E}^{t}$.}
\begin{algorithmic}[1]
\STATE $\mathbf H_b^{t+1}=\mathbf C_b^{t+1}=\mathbf H_f^{t-T_t}=\mathbf C_f^{t-T_t}=\mathbf 0$.
\FOR{$k=t$ to $t-T_t+1$}
    \STATE $\mathbf{H}_b^{k},\mathbf{C}_b^{k}  =  \mbox{ConvLSTM}(\mathbf{H}_b^{k+1},\mathbf{C}_b^{k+1}, \mathbf{E}^{k}, \theta_l^b)$.
\ENDFOR
\FOR{$k=t-T_t+1$ to $t$}
    \STATE $\mathbf{H}_f^{k},\mathbf{C}_f^{k}  =  \mbox{ConvLSTM}(\mathbf{H}_f^{k-1},\mathbf{C}_f^{k-1}, \mathbf{H}_b^{k}, \theta_l^f)$.
\ENDFOR
\STATE $\mathbf{\hat E}^{t} \gets \mathbf{H}_f^{t}$.
\end{algorithmic}
\label{alg:convlstm-cb}
\end{algorithm}

\begin{algorithm}[t]
\caption{Fused Bidirectional ConvLSTM.}
\fontsize{8pt}{10pt}
\selectfont
\KwIn{ Features: $\{\mathbf{E}^{k}|k=t,\cdots, t-T_t+1\}$; \\
 ConvLSTM parameters: $\theta_l^b$, $\theta_l^f$. }
\KwOut{Enhanced feature: $\mathbf{\hat E}^{t}$.}
\begin{algorithmic}[1]
\STATE $\mathbf H_b^{t+1}=\mathbf C_b^{t+1}=\mathbf H_f^{t-T_t}=\mathbf C_f^{t-T_t}=\mathbf 0$.
\FOR{$k=t$ to $t-T_t+1$}
    \STATE $\mathbf{H}_b^{k},\mathbf{C}_b^{k}  =  \mbox{ConvLSTM}(\mathbf{H}_b^{k+1},\mathbf{C}_b^{k+1}, \mathbf{E}^{k}, \theta_l^b)$.
\ENDFOR
\FOR{$k=t-T_t+1$ to $t$}
    \STATE $\mathbf{H}_f^{k},\mathbf{C}_f^{k}  =  \mbox{ConvLSTM}(\mathbf{H}_f^{k-1},\mathbf{C}_f^{k-1}, [\mathbf{H}_b^{k},\mathbf{E}^{k}], \theta_l^f)$.
\ENDFOR
\STATE $\mathbf{\hat E}^{t} \gets \mathbf{H}_f^{t}$.
\end{algorithmic}
\label{alg:convlstm-fb}
\end{algorithm}
\begin{algorithm}[t]
\caption{One training step of our self-enhanced convolutional network.}
\fontsize{8pt}{10pt}
\selectfont
\KwIn{LR and ground-truth image sequences:
$\mathcal X=\{\mathbf{X}^t\}$ and $\mathcal Y=\{\mathbf{Y}^t\}$ where $t=1,\cdots,N$; initialized network parameters: $\theta$.}
\KwOut{Optimized network parameters: $\theta$.}
\begin{algorithmic}[1]
\STATE $\mathrm L \gets 0$, $\mathcal E \gets \varnothing$, $\mathcal Y \gets \varnothing$.
\FOR{$t=1$ to $N$}
    \STATE Fetch LR frames from $\mathcal X$: $\mathbb X^t=\{\mathbf{X}^{k}| k=t-T_1,\cdots,t+T_1\}$.\newline If $k\leq0$, $\mathbf X^{k}=\mathbf X^1$; if $k>N$, $\mathbf X^{k}=\mathbf X^N$.
    \STATE Compute motion fields from frame $\mathbf X^t$ to all frames in $\mathbb X^t$: $\{\mathbf{F}^{t\rightarrow k}| k=t-T_1,\cdots,t+T_1; \; \mathbf{F}^{t\rightarrow t}=\mathbf 0\}$.
    \STATE Warp $\mathbb X^t$ to $\mathbb {\tilde X}^t=\{\mathbf{X}^{k \rightarrow t}\}$ using above motion fields.
    \STATE Compute the initial SR result $\mathbf{\hat Y}^t$ using the local frame fusion network in Section~\ref{sec:lffn} with input $\mathbb {\tilde X}^t$.
    \STATE Fetch super-resolved results from $\mathcal Y$: $\mathbb Y^t=\{\mathbf Y^{k}|k=t-1,\cdots,t-T_2;\}$. If $k\leq0$, $\mathbf Y^{k}=\mathbf 0$.
    \STATE Warp frames in $\mathbb Y^t$ to frame $t$ using bi-linearly upsampled motion fields, forming $\mathbb{\tilde Y}^t=\{\mathbf Y^{k\rightarrow t}\}$.
    \STATE Input $\mathbb{\tilde Y}^t$ and $\mathbf{\hat X}^t$ into the encoder of recurrent frame fusion module in Section~\ref{sec:erffn} and extract feature maps $\mathbf E_1^t$, $\mathbf E_2^t$ and $\mathbf E_3^t$.
    \STATE Fetch feature maps from $\mathcal E$: $\mathbb E^t=\{\mathbf E_3^{k}|k=t-1,\cdots,t-T_t+1;\;T_t=\min(t,T_3+1)\}$.
    \STATE Sequentially encode $\{\mathbf E_3^t\} \bigcup \mathbb E^t  $ into  $\mathbf{\hat E}_3^t$ using ConvLSTM cells.
    \STATE Feed $\mathbf E_1^t$, $\mathbf E_2^t$ and $\mathbf{\hat E}_3^t$ into the decoder, resulting to $\mathbf Y^t$.
    \STATE Compute $L_f^t$, $L_l^t$, and $L_e^t$ according to (\ref{eq:loss-flow}), (\ref{eq:loss-lffn}) and (\ref{eq:loss-erffn}) respectively.
    \STATE $L\gets L+L_e^t+L_l^t+\gamma L_f^t$. 
    \STATE Append $\mathbf E_3^t$ and $\mathbf Y^t$ into  $\mathcal E$ and $\mathcal Y$ respectively.
\ENDFOR
\STATE $L \gets L/N$; update $\theta$ using Adam.
\end{algorithmic}
\label{alg:train}
\end{algorithm}

\subsubsection{Encoder-decoder Framework}
Super-resolving HR images from LR images requires recovering both accurate global appearances and visual friendly details such as textures and sharp edges. Inspired by~\cite{su2017deep}, we adopt an encoder-decoder framework with skip connections to extract multi-scale convolutional features. Detailed network architecture is presented in right part of Fig. \ref{fig:model} (a). The encoder module extracts three scales of features using convolution layers, ReLUs~\cite{nair2010rectified} and residual blocks~\cite{he2016deep} from the input. Features in the first two scales are responsible for restoring details. They are forwarded to the decoder via skip connections. Compared to the first two scales, 4 extra residual blocks are applied in the third scale for sake of enlarging receptive field and producing deeper features. Denote feature map in the $i$-th scale as $\mathbf E_i^t$. The decoder possesses almost symmetric architecture with the encoder, generating the final restored result $\mathbf{Y}^t$  through refining the initial super-resolved result $\mathbf{\hat Y}^t$.
It should be noted in particular that: 1) The input of our encoder-decoder model is formed by concatenating $\mathbf{\hat Y}^t$ and the aligned super-resolved images of previous frames; 2) Feature maps of past frames in the third scale $\{\mathbf E_3^{k} | k<t\}$ are accumulated to enhance $\mathbf E_3^t$ to $\mathbf{\hat E}_3^t$ via sequential encoding strategies which will be introduced in Section~\ref{sec:sfe}.
The following mean square error loss function is exploited to train the above model,
\begin{equation}\label{eq:loss-erffn}
L_e^{t} = \frac{1}{cr^2wh} \|\mathbf{ Y}^t - \mathbf G^t\|_{2}^2.
\end{equation}

\subsubsection{Recurrent Frame Fusion}\label{sec:rff}
Because of the high dependency across frames in video, propagating super-resolved result of previous frame is helpful to infer the result of current frame in video super-resolution~\cite{sajjadi2018frame}. Here we take advantage of multiple previous frames $\{\mathbf Y^{k}|k=t-1,\cdots,t-T_2\}$ when predicting HR image of frame $t$. Each previous frame $\mathbf{Y}^{k}$ is firstly aligned to frame $t$ using the bi-linearly interpolated optical flow field $\mathbf F^{t\rightarrow k}$. Suppose the aligned result be $\mathbf{Y}^{k \rightarrow t}$. Afterwards $\{\mathbf{Y}^{k \rightarrow t}|k=t-1,\cdots,t-T_2\}$ are concatenated with $\mathbf{\hat Y}^t$ and then fed into the encoder. When $T_2\leq T_1$, warping past HR frames does not entail much more computational cost as all optical flow fields have been calculated in Section \ref{sec:lffn}. The differences with~\cite{sajjadi2018frame} are that multiple previous frames are utilized and no space-to-depth transformation is required to convert the HR image into a tensor with same spatial size as the LR image.

\subsubsection{Sequential Feature Encoding}\label{sec:sfe}
Feature-level temporal information benefits facial video hallucination from the following perspectives: facial motions could be used to infer future frames which is paramount to restore lost appearance information in future frames; motions in most regions of talking faces are usually not severe, making spatial dependencies in high-level feature maps could be easily obtained for ensuring content coherence in restored videos. Considering the above two points, we adopt ConvLSTM~\cite{xingjian2015convolutional} to extract temporal information for enhancing the feature representation of current frame. 
Given a sequence of input features $\{\mathbf E^k\}$. We summarize the formulation of ConvLSTM as follows,
\begin{equation}
\label{eq:backconvlstm} \mathbf{H}^{k},\mathbf{C}^{k}  =  \mbox{ConvLSTM}(\mathbf{H}^{k-1},\mathbf{C}^{k-1}, \mathbf{E}^{k}, \theta_l),
\end{equation}
where $\theta_l$ contains all the weights and biases of convolution kernels in the ConvLSTM cell. $\mathbf{H}^{k}$ and $\mathbf{C}^{k}$ is hidden state and cell output respectively. $\mathbf{H}^{0}=\mathbf{C}^{0}=\mathbf 0$.
Detailed computation steps of ConvLSTM are given below.
\begin{itemize}
\item[1.] $\mathbf A_i^k=\varsigma(\mathbf E^k \ast \mathbf W_{ei}+\mathbf H^{k-1} \ast \mathbf W_{hi}+\mathbf b_i)$;
\item[2.] $\mathbf A_f^k=\varsigma(\mathbf E^k \ast \mathbf W_{ef}+\mathbf H^{k-1}\ast \mathbf W_{hf}+\mathbf b_f)$;
\item[3.] $\mathbf A_g^k=\tanh(\mathbf E^k\ast\mathbf W_{eg}+\mathbf H^{k-1} \ast \mathbf W_{hg}+\mathbf b_g)$;
\item[4.] $\mathbf C^k=\mathbf{A}_f^k\circ \mathbf{C}^{k-1}+\mathbf A_i^k\circ \mathbf A_g^k $;
\item[5.] $\mathbf A_o^k=\varsigma(\mathbf E^k \ast\mathbf W_{io}+\mathbf H^{k-1} \ast \mathbf W_{ho}+\mathbf b_o)$;
\item[6.] $\mathbf H^k=\mathbf A_o^k \circ \tanh(\mathbf C^k)$,
\end{itemize}
where $\varsigma(\cdot)$ is the Sigmoid function. $\mathbf{W}$-s and $\mathbf b$-s are the weights and biases of convolution kernels with size of $3\times3$. `$\circ$' represents the Hadamard product. $\mathbf A_i^k$, $\mathbf A_f^k$ and $\mathbf A_o^k$ represent input, forget and output gate for the $k$-th data sample, respectively.

We provide three strategies to encode collected features based on ConvLSTM units: one-way ConvLSTM, cascaded bidirectional ConvLSTM and fused bidirectional ConvLSTM. 

\begin{table*}[t]
\small
\caption{Datasets used in facial and generic video super-resolution.}\label{tab:dset}
\centering
\fontsize{8pt}{10pt}
\selectfont
\setlength\tabcolsep{3pt}
\begin{tabular}{ c||c|c|c|c||c|c|c|c||c|c|c|c } \specialrule{.1em}{0em}{0em}
\multicolumn{1}{c||}{\multirow{2}{*}{}} &  \multicolumn{4}{c||}{training} &  \multicolumn{4}{c||}{validation} &  \multicolumn{4}{c}{testing} \\ \cline{2-13}
 &persons & sequences & frames & size & persons  & sequences &  frames & size & persons  & sequences & frames & size \\ \hline
VoxCeleb
&922 &140,334 &1,102,792 & $280\times280$ &4 &14 &1,209 &$256\times256$ &80 &334 &28,636 & $256\times256$ \\
RAVDESS
&0    &0   &0 &- &0 &0  &0 &- &24 &96 &10,013 &$256\times256$ \\ 
Harmonic
&- &7,456 &1,163,056 & $300\times300$ &- &4 &798 &$512\times512$ &-   &51 &9,335 &$512\times640$ \\
VID4
&- &- &- & - &- &- &- &- &-   &4 &171 &not fixed \\ \specialrule{.1em}{0em}{0em}
\end{tabular}
\end{table*}

\noindent \textbf{One-way ConvLSTM:} To capture long-term temporal information, we propagate the cell output $\mathbf C$ and hidden state $\mathbf H$ to next frame recurrently as shown in Fig.~\ref{fig:sqe} (a). Consequently at any frame $t$ all past features $\{\mathbf E_3^{k}|k<t\}$ are exploited to enhance $\mathbf E_3^t$,
\begin{equation}
\mathbf{H}^{t},\mathbf{C}^{t}  =  \mbox{ConvLSTM}(\mathbf{H}^{t-1},\mathbf{C}^{t-1}, \mathbf{E}_3^{t}, \theta_l).
\end{equation}
\noindent The result of the enhanced feature $\mathbf{\hat E}_3^{t}$ is $\mathbf{H}^{t}$ exactly. The computation procedure is concluded in Algorithm~\ref{alg:convlstm-one}.

\noindent \textbf{Cascaded Bidirectional ConvLSTM:}  According to~\cite{schuster1997bidirectional}, bidirectional RNN framework outperforms regular recurrent model with one-way pass. Thus we can devise a bidirectional ConvLSTM module as shown in Fig.~\ref{fig:sqe} (b). Here only $T_3$ past features at most should be considered, preventing the time and memory cost from increasing continuously as $t$ grows. Then one backward and forward passes are adopted to processing the ordered feature sequence $\mathbb E^t=\{\mathbf E_3^{k}| k=t,\cdots,t-T_t+1;\;T_t=\min(t,T_3+1)\}$,
\begin{align}
\label{eq:backpass} \mathbf{H}_b^{k},\mathbf{C}_b^{k}  &=  \mbox{ConvLSTM}(\mathbf{H}_b^{k+1},\mathbf{C}_b^{k+1}, \mathbf{E}_3^{k}, \theta_l^b); \\
\label{eq:forwpass} \mathbf{H}_f^{k},\mathbf{C}_f^{k}  &=  \mbox{ConvLSTM}(\mathbf{H}_f^{k-1},\mathbf{C}_f^{k-1}, \mathbf{H}_b^{k}, \theta_l^f).
\end{align}
\noindent $\mathbf{H}_b^{k}$, $\mathbf{C}_b^{k}$ and $\theta_l^b$ are the hidden state, cell output and parameter of the backward pass respectively while $\mathbf{H}_f^{k}$, $\mathbf{C}_f^{k}$ and $\theta_l^f$ represents the hidden state, cell output and parameter of the forward pass respectively. The final result $\mathbf{\hat E}_3^t$ is $\mathbf{H}_f^{1}$. $\mathbf{H}_b^{t+1}=\mathbf{C}_b^{t+1}=\mathbf 0$. $\mathbf{H}_f^{t-T_t}=\mathbf{C}_f^{t-T_t}=\mathbf 0$. The computation procedure is summarized in Algorithm~\ref{alg:convlstm-cb}.

\noindent \textbf{Fused Bidirectional ConvLSTM:} To prevent loss of forward motion information, we devise another sequential feature encoding strategy as shown in Fig.~\ref{fig:sqe} (c). Feature maps are not only fed into the ConvLSTM cell of the backward pass, but also constitute proportion of the input of the forward ConvLSTM cell as shown in Algorithm \ref{alg:convlstm-fb}. The backward pass is the same as (\ref{eq:backpass}) while the forward pass (\ref{eq:forwpass}) is replaced with the following procedure,
\begin{equation}
\mathbf{H}_f^{k},\mathbf{C}_f^{k}  =  \mbox{ConvLSTM}(\mathbf{H}_f^{k+1},\mathbf{C}_f^{k+1}, [\mathbf{H}_b^{k},\mathbf{E}_3^{k}], \theta_l^f).
\end{equation}

\subsubsection{Self-learned Attention}
\begin{figure}[t]
\centering
\includegraphics[width=\columnwidth]{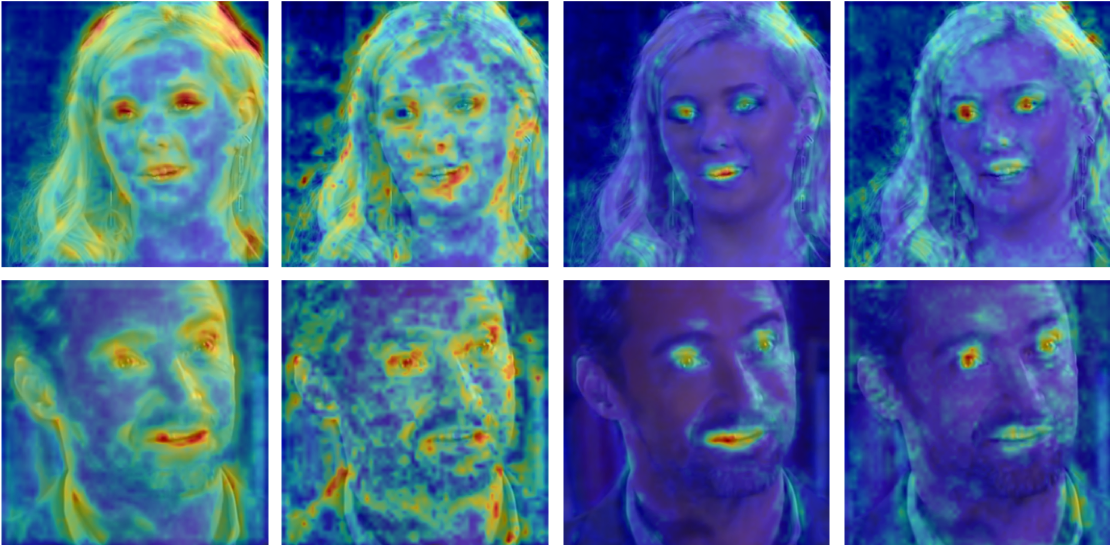}
\caption{Visualizations of self-learned spatial attention maps from the 1st, 2nd, 4th and 6th residual blocks.}
\label{fig:attention}
\end{figure}

Spatial attention is significant in facial image hallucination as faces consist of specific components. Inspired by \cite{hu2018squeeze}, we integrate a spatial attention mechanism into the residual block as shown in Fig. \ref{fig:model} (d). Two convolution layers and one ReLU layer are used to produce a spatial attention map, which is subsequently applied to suppress activations of pixels with low attention values. This attention mechanism enables every residual block to emphasize particular regions. Examples of our self-learned attention maps are presented in Fig. \ref{fig:attention}.

\subsection{Network Training} \label{sec:train}
 Summing up (\ref{eq:loss-flow}) (\ref{eq:loss-lffn}) and (\ref{eq:loss-erffn}), we can obtain the overall training loss,
\begin{equation} \label{eq:loss}
L=\frac{1}{N}\sum_{t=1}^N (L_e^t+L_l^t+\gamma L_f^{t}),
\end{equation}
where $\gamma$ is a constant. The loss function (\ref{eq:loss}) is optimized using Adam~\cite{kingma2014adam} with learning rate of $10^{-4}$. One step of optimization is illustrated in Algorithm~\ref{alg:train}. During the inference stage, only those super-resolved images and features required in next frame are preserved at the end of each frame. 

\begin{figure*}[t]
\centering
\begin{tabular}{p{0.12\linewidth}<{\centering}p{0.12\linewidth}<{\centering}
p{0.12\linewidth}<{\centering}p{0.12\linewidth}<{\centering}
p{0.12\linewidth}<{\centering}p{0.12\linewidth}<{\centering}
p{0.12\linewidth}<{\centering}}
 Bicubic & LapSRNet & FSRNet  & SPMC & FRVSR & Ours & GT \\
\end{tabular}
\begin{minipage}{\linewidth}
\centering
\includegraphics[width=0.135\textwidth]{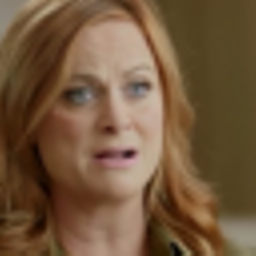}
\includegraphics[width=0.135\textwidth]{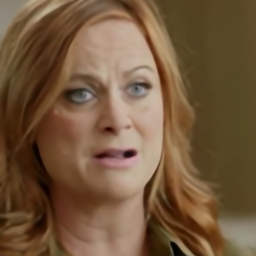}
\includegraphics[width=0.135\textwidth]{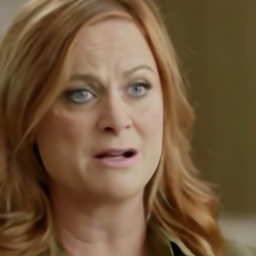}
\includegraphics[width=0.135\textwidth]{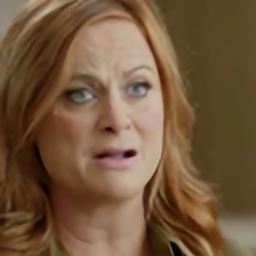}
\includegraphics[width=0.135\textwidth]{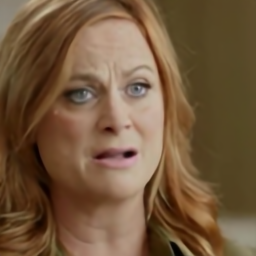}
\includegraphics[width=0.135\textwidth]{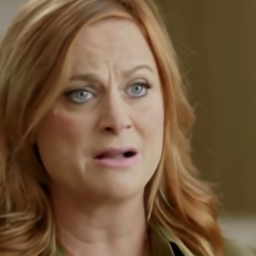}
\includegraphics[width=0.135\textwidth]{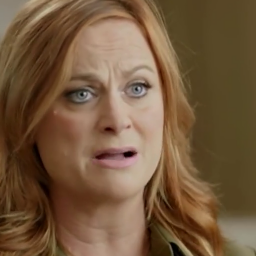}
\end{minipage}\\ \vspace{1mm}
\begin{minipage}{\linewidth}
\centering
\includegraphics[width=0.135\textwidth]{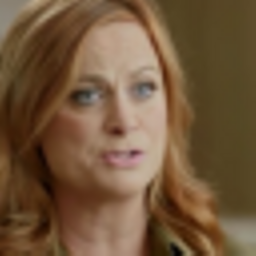}
\includegraphics[width=0.135\textwidth]{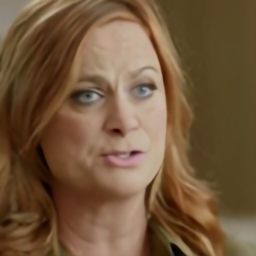}
\includegraphics[width=0.135\textwidth]{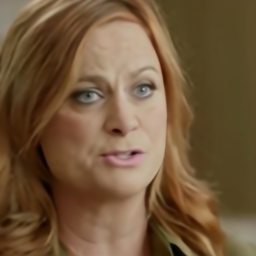}
\includegraphics[width=0.135\textwidth]{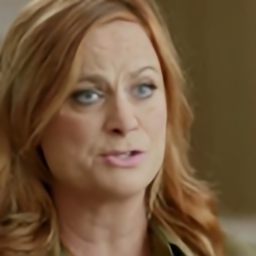}
\includegraphics[width=0.135\textwidth]{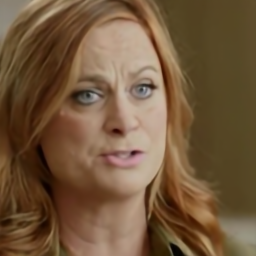}
\includegraphics[width=0.135\textwidth]{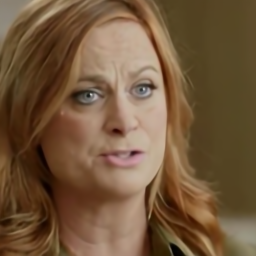}
\includegraphics[width=0.135\textwidth]{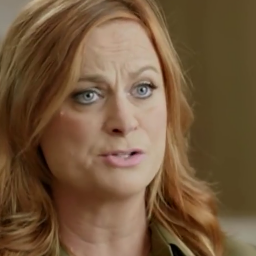}
\end{minipage}\\ \vspace{1mm}
\begin{minipage}{\linewidth}
\centering
\includegraphics[width=0.135\textwidth]{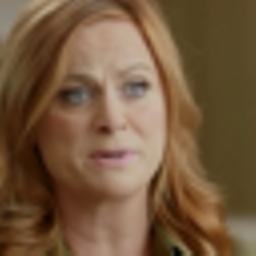}
\includegraphics[width=0.135\textwidth]{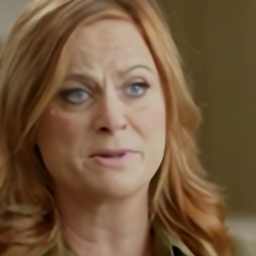}
\includegraphics[width=0.135\textwidth]{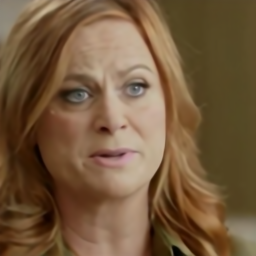}
\includegraphics[width=0.135\textwidth]{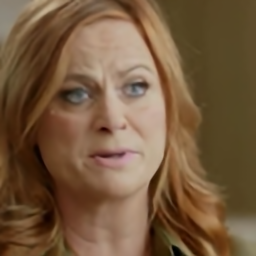}
\includegraphics[width=0.135\textwidth]{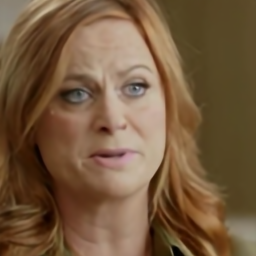}
\includegraphics[width=0.135\textwidth]{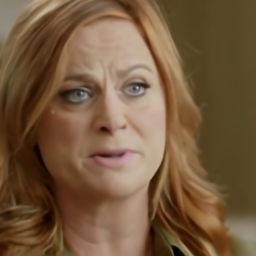}
\includegraphics[width=0.135\textwidth]{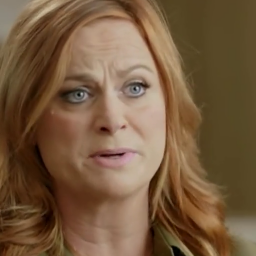}
\end{minipage}\\
\begin{tabular}{p{0.12\linewidth}<{\centering}p{0.12\linewidth}<{\centering}
p{0.12\linewidth}<{\centering}p{0.12\linewidth}<{\centering}
p{0.12\linewidth}<{\centering}p{0.12\linewidth}<{\centering}
p{0.12\linewidth}<{\centering}}
 31.40/0.8869 & 35.04/0.9397 &35.19/0.9425 & 35.36/0.9441 & 36.96/0.9596 & 37.88/0.9660  & PSNR/SSIM$_{\mathrm{vh}}$ \\
\end{tabular}
\begin{minipage}{\linewidth}
\centering
\includegraphics[width=0.135\textwidth]{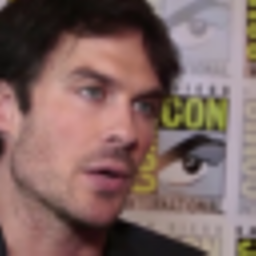}
\includegraphics[width=0.135\textwidth]{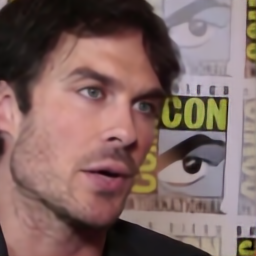}
\includegraphics[width=0.135\textwidth]{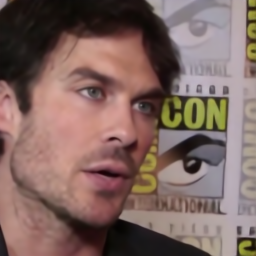}
\includegraphics[width=0.135\textwidth]{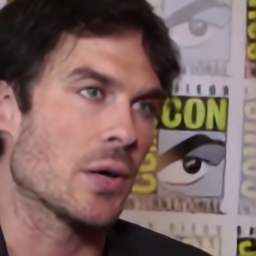}
\includegraphics[width=0.135\textwidth]{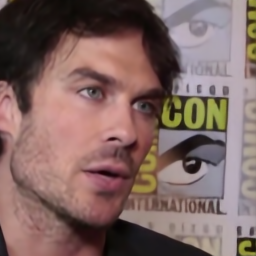}
\includegraphics[width=0.135\textwidth]{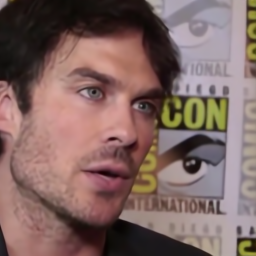}
\includegraphics[width=0.135\textwidth]{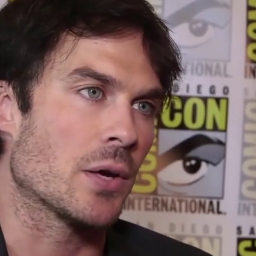}
\end{minipage}\\ \vspace{1mm}
\begin{minipage}{\linewidth}
\centering
\includegraphics[width=0.135\textwidth]{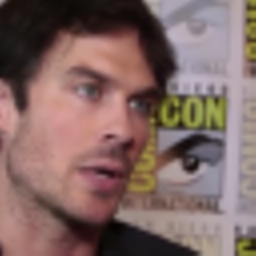}
\includegraphics[width=0.135\textwidth]{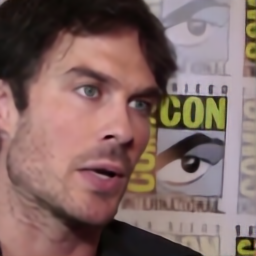}
\includegraphics[width=0.135\textwidth]{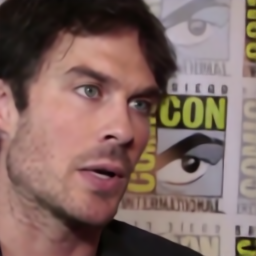}
\includegraphics[width=0.135\textwidth]{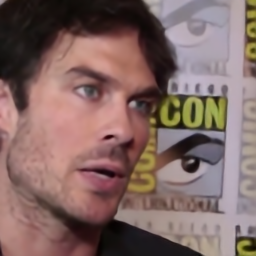}
\includegraphics[width=0.135\textwidth]{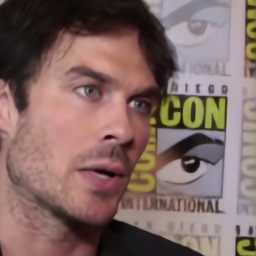}
\includegraphics[width=0.135\textwidth]{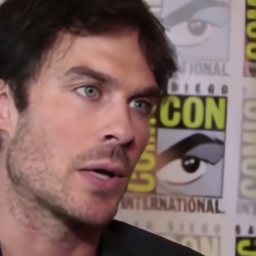}
\includegraphics[width=0.135\textwidth]{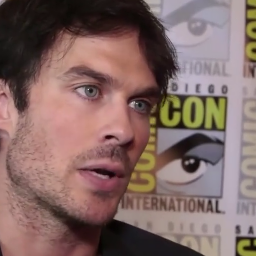}
\end{minipage}\\ \vspace{1mm}
\begin{minipage}{\linewidth}
\centering
\includegraphics[width=0.135\textwidth]{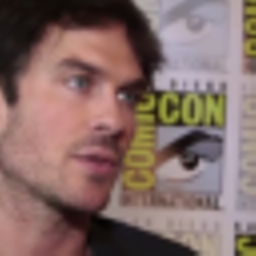}
\includegraphics[width=0.135\textwidth]{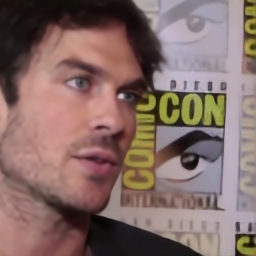}
\includegraphics[width=0.135\textwidth]{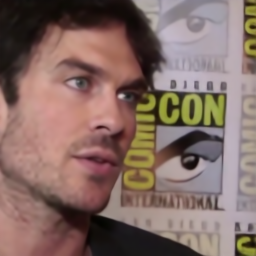}
\includegraphics[width=0.135\textwidth]{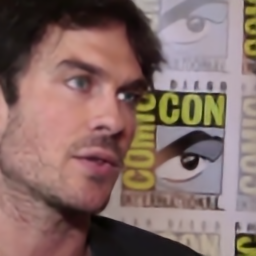}
\includegraphics[width=0.135\textwidth]{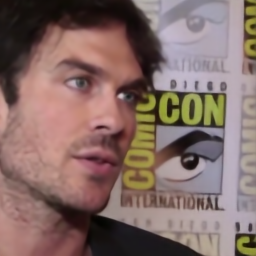}
\includegraphics[width=0.135\textwidth]{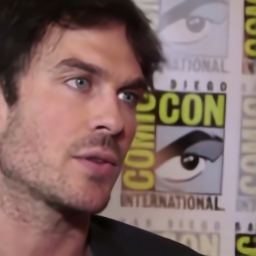}
\includegraphics[width=0.135\textwidth]{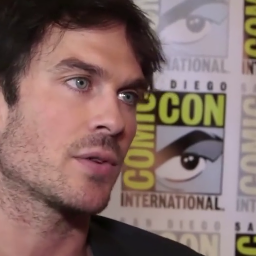}
\end{minipage}\\
\begin{tabular}{p{0.12\linewidth}<{\centering}p{0.12\linewidth}<{\centering}
p{0.12\linewidth}<{\centering}p{0.12\linewidth}<{\centering}
p{0.12\linewidth}<{\centering}p{0.12\linewidth}<{\centering}
p{0.12\linewidth}<{\centering}}
 28.25/0.8481 &31.78/0.9160 &32.79/0.9279 & 32.39/0.9231 & 34.18/0.9450 & 35.86/0.9590  & PSNR/SSIM$_{\mathrm{vh}}$ \\
\end{tabular}
\caption{Comparison of super-resolution algorithms in two facial image sequences. The predicted SR images from our method (`Ours') are closer to the ground truth than other algorithms.}
\label{fig:comp}
\end{figure*}
\begin{figure*}[t]
\centering
\begin{tabular}{p{0.12\linewidth}<{\centering}p{0.12\linewidth}<{\centering}
p{0.12\linewidth}<{\centering}p{0.12\linewidth}<{\centering}
p{0.12\linewidth}<{\centering}p{0.12\linewidth}<{\centering}
p{0.12\linewidth}<{\centering}}
 Bicubic & BRCN & SPMC & FRVSR & VSR-DUF & Ours & GT \\
\end{tabular}
\begin{minipage}{\linewidth}
\centering
\includegraphics[width=0.135\textwidth]{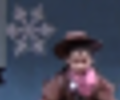}
\includegraphics[width=0.135\textwidth]{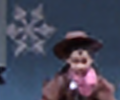}
\includegraphics[width=0.135\textwidth]{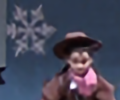}
\includegraphics[width=0.135\textwidth]{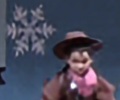}
\includegraphics[width=0.135\textwidth]{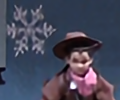}
\includegraphics[width=0.135\textwidth]{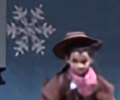}
\includegraphics[width=0.135\textwidth]{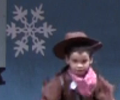}
\end{minipage}\\
\begin{tabular}{p{0.12\linewidth}<{\centering}p{0.12\linewidth}<{\centering}
p{0.12\linewidth}<{\centering}p{0.12\linewidth}<{\centering}
p{0.12\linewidth}<{\centering}p{0.12\linewidth}<{\centering}
p{0.12\linewidth}<{\centering}}
 27.92/0.8526 & 29.09/0.8800 & 30.71/0.9175 & 30.41/0.9179 & 29.08/0.9118 & 30.82/0.9311  & PSNR/SSIM \\
\end{tabular}
\begin{minipage}{\linewidth}
\centering
\includegraphics[width=0.135\textwidth]{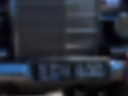}
\includegraphics[width=0.135\textwidth]{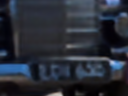}
\includegraphics[width=0.135\textwidth]{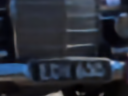}
\includegraphics[width=0.135\textwidth]{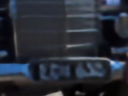}
\includegraphics[width=0.135\textwidth]{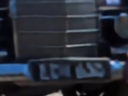}
\includegraphics[width=0.135\textwidth]{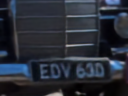}
\includegraphics[width=0.135\textwidth]{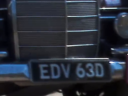}
\end{minipage}\\
\begin{tabular}{p{0.12\linewidth}<{\centering}p{0.12\linewidth}<{\centering}
p{0.12\linewidth}<{\centering}p{0.12\linewidth}<{\centering}
p{0.12\linewidth}<{\centering}p{0.12\linewidth}<{\centering}
p{0.12\linewidth}<{\centering}}
 23.96/0.6977 & 24.93/0.7537 & 25.71/0.8009 & 25.83/0.8045 & 27.37/0.8682 & 29.33/0.9266  & PSNR/SSIM \\
\end{tabular}
\begin{minipage}{\linewidth}
\centering
\includegraphics[width=0.135\textwidth]{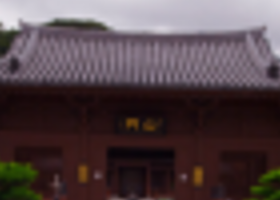}
\includegraphics[width=0.135\textwidth]{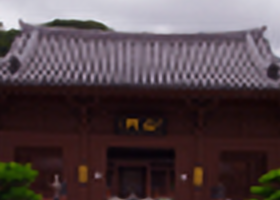}
\includegraphics[width=0.135\textwidth]{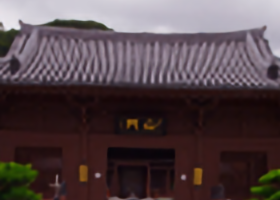}
\includegraphics[width=0.135\textwidth]{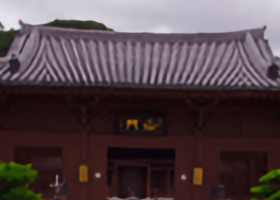}
\includegraphics[width=0.135\textwidth]{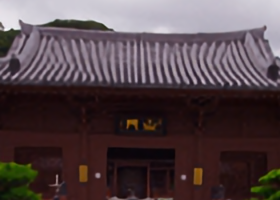}
\includegraphics[width=0.135\textwidth]{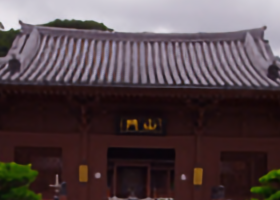}
\includegraphics[width=0.135\textwidth]{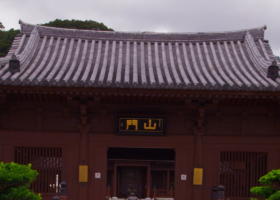}
\end{minipage}\\
\begin{tabular}{p{0.12\linewidth}<{\centering}p{0.12\linewidth}<{\centering}
p{0.12\linewidth}<{\centering}p{0.12\linewidth}<{\centering}
p{0.12\linewidth}<{\centering}p{0.12\linewidth}<{\centering}
p{0.12\linewidth}<{\centering}}
 25.20/0.7506 & 26.33/0.8056 & 26.17/0.8159 & 24.06/0.7789 & 26.26/0.8455 & 31.21/0.9152  & PSNR/SSIM \\
\end{tabular}
\caption{Comparison of super-resolution algorithms in generic video super-resolution. }
\label{fig:compvsr}
\end{figure*}

\begin{figure*}[t]
\centering
\begin{tabular}{p{0.12\linewidth}<{\centering}p{0.12\linewidth}<{\centering}
p{0.12\linewidth}<{\centering}p{0.12\linewidth}<{\centering}
p{0.12\linewidth}<{\centering}p{0.12\linewidth}<{\centering}
p{0.12\linewidth}<{\centering}}
 \scriptsize Bicubic & \scriptsize w.o./\emph{rff}, w.o./\emph{sfe}  & \scriptsize w.o./\emph{rff}, cascaded &\scriptsize w./\emph{rff}, w.o./\emph{sfe} & \scriptsize w./\emph{rff}, cascaded & \scriptsize w./\emph{rff}, fused &\scriptsize GT \\
\end{tabular}
\begin{minipage}{\linewidth}
\centering
\includegraphics[width=0.135\textwidth]{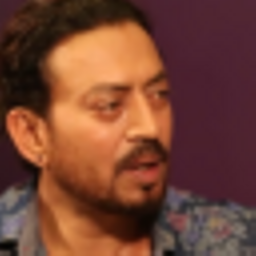}
\includegraphics[width=0.135\textwidth]{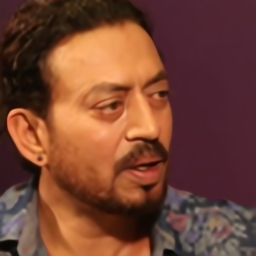}
\includegraphics[width=0.135\textwidth]{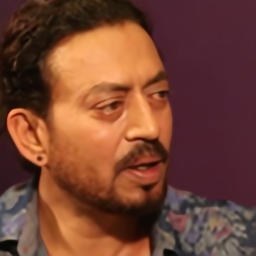}
\includegraphics[width=0.135\textwidth]{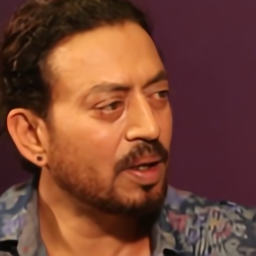}
\includegraphics[width=0.135\textwidth]{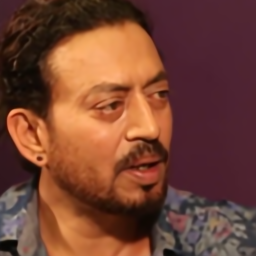}
\includegraphics[width=0.135\textwidth]{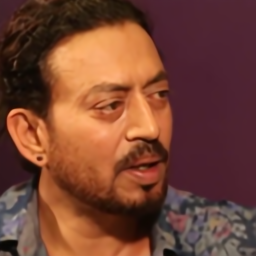}
\includegraphics[width=0.135\textwidth]{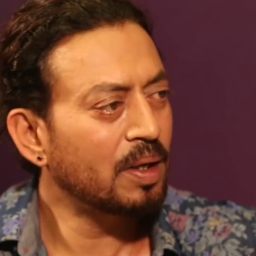}
\end{minipage}\\ \vspace{1mm}
\begin{minipage}{\linewidth}
\centering
\includegraphics[width=0.135\textwidth]{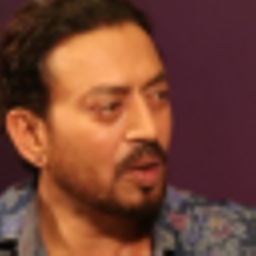}
\includegraphics[width=0.135\textwidth]{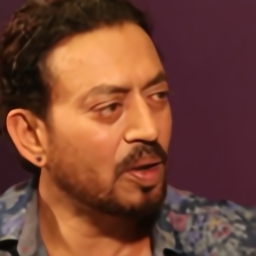}
\includegraphics[width=0.135\textwidth]{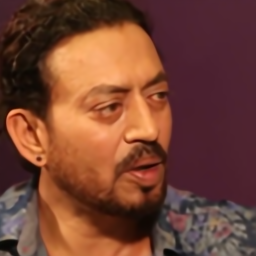}
\includegraphics[width=0.135\textwidth]{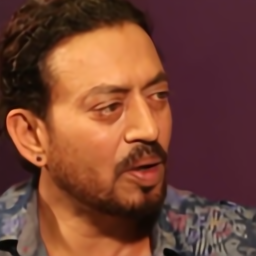}
\includegraphics[width=0.135\textwidth]{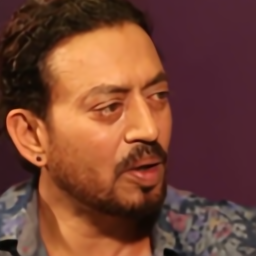}
\includegraphics[width=0.135\textwidth]{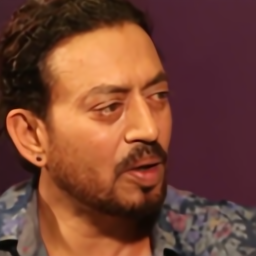}
\includegraphics[width=0.135\textwidth]{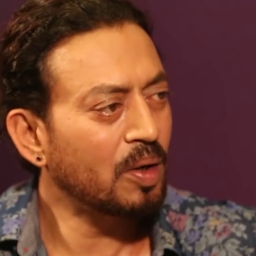}
\end{minipage}\\ \vspace{1mm}
\begin{minipage}{\linewidth}
\centering
\includegraphics[width=0.135\textwidth]{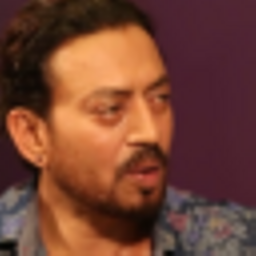}
\includegraphics[width=0.135\textwidth]{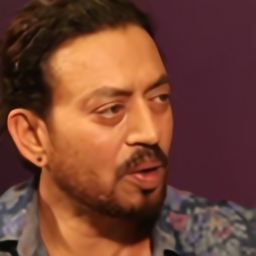}
\includegraphics[width=0.135\textwidth]{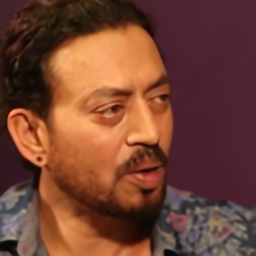}
\includegraphics[width=0.135\textwidth]{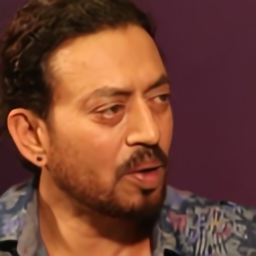}
\includegraphics[width=0.135\textwidth]{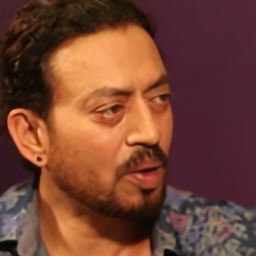}
\includegraphics[width=0.135\textwidth]{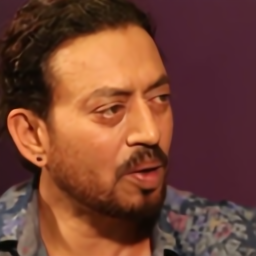}
\includegraphics[width=0.135\textwidth]{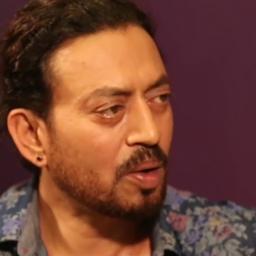}
\end{minipage}\\ \vspace{1mm}
\begin{tabular}{p{0.12\linewidth}<{\centering}p{0.12\linewidth}<{\centering}
p{0.12\linewidth}<{\centering}p{0.12\linewidth}<{\centering}
p{0.12\linewidth}<{\centering}p{0.12\linewidth}<{\centering}
p{0.12\linewidth}<{\centering}}
 31.41/0.8973 & 36.84/0.9581 & 37.57/0.9650 & 35.78/0.9611 & 38.58/0.9711 & 38.63/0.9715  & PSNR/SSIM \\
\end{tabular}
\caption{Comparison of different variants of our method. `w.o.' represents `without' and `w.' represents `with'. `cascaded' means cascaded bidirectional ConvLSTM is adopted for \emph{sfe} strategy while `fused' means fused bidirectional ConvLSTM is applied.}
\label{fig:compinner}
\end{figure*}
\section{Experiments}

\subsection{Dataset}
\label{sec:data}
Two facial video datasets, VoxCeleb~\cite{Nagrani17} and RAVDESS~\cite{livingstone2012ravdess}, are used to validate the performance of our method. In addition, we also use generic single-shot videos from Harmonic and VID4~\cite{liu2014bayesian} to test deep video super-resolution methods. Table~\ref{tab:dset} presents the split of training, validation and testing sets.
\begin{itemize}
\item[(1)]
The \textbf{VoxCeleb} dataset contains over 100,000 utterances by 1,251 celebrities, providing sequences of tracked faces in the form of bounding boxes. We select out 140,334 sequences of face images with high quality. For each sequence, we compute a box enclosing the faces from all frames and use it to crop face images from the original video. All face images are resized to $280\times 280$. Only the central $256\times256$ region is used in validating and testing.
\item[(2)]
The \textbf{RAVDESS} dataset encloses 2,452 sequences captured from 24 persons speaking/singing with various expressions and motions. We choose 4 sequences for each person forming the other testing set of facial video hallucination.
\item[(3)]
The \textbf{Harmonic} dataset includes 18 videos captured from natural scenes containing buildings, birds, animals, etc. Sequences without scene switching are manually selected to serve as our single-shot video super-resolution dataset. Images are resized to $540\times960$. For training set we uniformly sample 8 sequences of $300\times300$ images from every video clip along horizontal axis with stride of 220 and vertical axis with stride of 240. Validation set is generated via cropping the centering $512\times512$ regions. Testing images with size of $512\times640$ are also cropped out in the center. 
\item[(4)]
The \textbf{VID4} dataset containing 4 sequences (`calendar', `city', `foliage' and `walk') of images has been widely used to validate video SR methods.
\end{itemize}

Input LR images are synthesized through blurring HR images with a Gaussian kernel (standard deviation of 1.5), and then downscaling them via sampling 1 pixel out of every 4 pixels in each dimension. The following strategies are adopted for data augmentation during the training stage: 
\begin{itemize}
\item[1)] Random shuffle is utilized to reorganize the order of image sequences in each epoch;
\item[2)] $N$ consecutive frames are selected from each sequence in the batch via starting at a random position and sampling one frame out of every $l$ frames where $l$ is random integer within $[1,2]$ for facial videos and $[1,4]$ for general videos;  
\item[3)] $256\times256$ patches are randomly cropped from HR images of these frames, serving as ground-truth; 
\item[4)] The chronologically order of the selected frames is reversed randomly; 
\item[5)] Images are randomly flipped horizontally. 
\end{itemize}

\noindent \textbf{Parameters \& Training Settings} $N$, $\alpha$ and $\gamma$ is set as 8, 0.01 and 0.1 respectively. According to the discussion in~\cite{caballero2017real}, we fix the value of $T_1$ as 2 in this paper. Without specification, $T_2$ and $T_3$ is set to 2 and 6 respectively; 8 residual blocks are adopted. In the recurrent frame fusion part, the error back propagation to the optical flow module is cut off to alleviate the instability in training. The local frame fusion network is pretrained for $2\times10^5$ iterations. Then the overall model is trained for the other $1.5\times10^5$ steps. For the generic video SR task, models are additionally finetuned for $5\times10^4$ iterations using sequences of $512\times512$ ground-truth images randomly cropped from the original $540\times960$ HR images. Batch size in each training iteration is set as 4. We periodically (every 200 iterations) test the model in the validation set. The version with the best performance is regarded as the final model. 4 TITAN Xp 12GB GPUs are utilized for training. Network parameters are initialized by default in PyTorch.

\noindent \textbf{Abbreviations} For conciseness, we use the following abbreviations to mark variants/settings of our method: `SECNet', self-enhanced convolutional network as shown in Fig. \ref{fig:model}; `LFFNet', local frame fusion network as described in Section \ref{sec:lffn}; `ERFFNet', network formed from the enhanced frame fusion module (Section \ref{sec:erffn}) through using bicubicly upsampled LR images and super-resolved images of previous frames as inputs; `CLSTM', ConvLSTM; `BCLSTM', bi-directional ConvLSTM; `\emph{rff}', recurrent frame fusion; `\emph{sfe}', sequential feature encoding.

\begin{table*}[t]
\caption{Performance of facial video hallucination on the testing sets of VoxCeleb and RAVDESS. `\#Parameters' indicates number of trainable parameters in each SR model. `FPS' indicates number of frames processed by each method per second.}\label{tab:facesr}
\centering
\fontsize{8pt}{10pt}
\selectfont
\begin{tabular}{ l||l|l|c||l|l|c||r||r }\specialrule{.1em}{0em}{0em}
\multicolumn{1}{c||}{\multirow{2}{*}{}} &  \multicolumn{3}{c||}{VoxCeleb} & \multicolumn{3}{c||}{RAVDESS} 
&\multirow{2}{*}{\#Parameters} &\multirow{2}{*}{FPS}  \\ \cline{2-7}
& PSNR($F_t$/$F_p$) & SSIM$_{\mathrm{vh}}$($F_t$/$F_p$)& SSIM$_{\mathrm{vt}}$ & PSNR($F_t$/$F_p$)& SSIM$_{\mathrm{vh}}$($F_t$/$F_p$)& SSIM$_{\mathrm{vt}}$ & &  \\ \hline
Bicubic
&29.56(201/0.0) &0.8776(249/0.0) &0.8957 &26.88(317/0.0) &0.9036(200/0.0) &0.9100 & - & -  \\
GLN~\cite{tuzel2016global}
&32.71(83.0/0.0)  &0.9243(119/0.0) &0.9321 &32.60(42.3/0.0)  &0.9460(40.3/0.0) &0.9426 & 40,818,842 &745.2 \\
WaveletSRNet~\cite{huang2017wavelet}
&32.75(84.0/0.0)  &0.9257(117/0.0) &0.9333 &32.32(54.4/0.0) &0.9408(62.0/0.0) &0.9400 & 51,353,968 &146.8 \\
\cite{bulat2018super}
&32.85(79.0/0.0)  &0.9263(113/0.0) &0.9337 &32.81(34.5/0.0) &0.9471(35.9/0.0) &0.9440 & 1,332,355 &194.9 \\
LapSRNet~\cite{lai2017deep}
&32.91(76.9/0.0)  &0.9281(107/0.0) &0.9351 &32.75(36.6/0.0) &0.9473(35.3/0.0) &0.9440 & 873,824   &418.0 \\
FSRNet~\cite{chen2018fsrnet}
&32.94(78.5/0.0)  &0.9303(101/0.0) &0.9369 &32.70(38.9/0.0) &0.9488(29.1/0.0) &0.9456 & 9,074,899 &75.8  \\
SRResCNN~\cite{ledig2017photo}
&33.00(75.0/0.0) &0.9275(111/0.0)   &0.9345 &32.61(42.6/0.0) &0.9489(28.3/0.0) &0.9459 & 1,549,335 &136.4 \\
SRResCNN-GAN~\cite{ledig2017photo}
&30.82(152/0.0)&0.8868(220/0.0)     &0.8983 &30.73(118/0.0) &0.9348(78.8/0.0) &0.9327 & - &- \\
SRResACNN~\cite{pathak2018efficient}
&33.35(61.6/0.0) &0.9319(94.6/0.0)  &0.9383 &33.23(18.7/0.0) &0.9508(20.7/0.0) &0.9475 & 1,557,688 &115.2 \\
SRResACNN-GAN~\cite{pathak2018efficient}
&32.23(97.9/0.0) &0.9167(139/0.0)   &0.9243 &32.18(56.3/0.0) &0.9437(46.0/0.0) &0.9400 & - &- \\
ESRCNN~\cite{wang2018esrgan}
&33.74(49.1/0.0) &0.9375(73.0/0.0)  &0.9433 &33.52(8.23/0.0) &0.9527(13.0/0.0) &0.9493 & 16,697,987 &17.7 \\ 
ESRCNN-GAN~\cite{wang2018esrgan}
&32.43(89.9/0.0) &0.9191(130/0.0)   &0.9262 &32.27(50.9/0.0) &0.9444(43.5/0.0) &0.9407 & - &- \\ \hline
BRCN~\cite{huang2015bidirectional}
&31.63(121/0.0)  &0.9091(166/0.0) &0.9200 &30.50(134/0.0) &0.9329(92.8/0.0) &0.9309 &90,828 & 295.6 \\
VESPCN~\cite{caballero2017real}
&32.62(85.7/0.0) &0.9266(110/0.0) &0.9331 &31.32(96.8/0.0) &0.9393(67.2/0.0) &0.9354 &109,528 & 109.0 \\
SPMC~\cite{tao2017detail}
&33.08(70.4/0.0) &0.9309(96.0/0.0)  &0.9372 &32.67(40.1/0.0) &0.9474(34.5/0.0) &0.9439 &1,731,363 & 49.1 \\
FRVSR~\cite{sajjadi2018frame}
&34.33(29.8/0.0) &0.9458(40.5/0.0) &0.9493 &33.26(17.7/0.0) &0.9515(18.1/0.0) &0.9475 &5,281,509 & 106.7\\
VSR-DUF$^*$~\cite{jo2018deep}
&33.63(52.0/0.0) &0.9455(41.4/0.0) &0.9483 &32.14(61.5/0.0) &0.9493(27.6/0.0) &0.9443 &5,821,952 & 9.3 \\\hline
LFFNet
&33.76(48.2/0.0) &0.9398(64.3/0.0) &0.9435 &32.74(37.9/0.0) &0.9484(31.4/0.0) &0.9436 &3,382,405 &105.5 \\
ERFFNet
&34.93(10.5/0.0) &0.9528(10.2/0.0) &0.9552 &33.49(9.50/0.0)  &0.9544(6.26/0.0)  &0.9503 &4,551,237 &28.8 \\
SECNet
&\textbf{35.26(0.00/1.0)} &\textbf{0.9550(0.00/1.0)} &\textbf{0.9572}
&\textbf{33.75(0.00/1.0)} &\textbf{0.9558(0.00/1.0)} &\textbf{0.9517} &5,334,792 &28.5 \\
\specialrule{.1em}{0em}{0em}
\end{tabular}
\end{table*}
\subsection{Evaluation Metrics}
PSNR and SSIM are employed to evaluate the performance of video SR methods. PSNR is computed using the mean squared error of image sequence $\{\mathbf Y^t\}$ in comparison to $\{\mathbf G^t\}$,
\begin{equation}
    \text{PNSR} =  \min(\log_{10}\frac{R}{\sqrt{\sum_{t=1}^N\|\mathbf Y^t - \mathbf G^t\|_2^2/(Ncr^2hw) }},100),
\end{equation}
where $R$ is the pixel range. $R$ is set to 1 as all images are normalized to [0,1].

SSIM is widely used for evaluating perceived quality of digital images. In this paper it is calculated over individual RGB images of videos. Given two patches $x$, $y$ from super-resolved and GT images respectively, the SSIM measure is calculated as follows,
\begin{equation} \label{eq:ssim}
\text{SSIM}(x,y) = \frac{(2 \mu_x \mu_y+c_1)( 2\sigma_{xy}+c_2)}{(\mu_x^2+\mu_y^2+c_1)(\sigma_x^2+\sigma_y^2+c_2)},
\end{equation}
where $c_1=(0.01R)^2$, $c_2=(0.03R)^2$. $\mu_x$, $\sigma_x$ and $\sigma_{xy}$ is the pixel average of $x$, standard deviation of $x$ and covariance between $x$ and $y$ respectively, 
\begin{eqnarray}
\mu_x & = & \sum_{i=-d}^d \sum_{j=-d}^d w_{ij}x_{ij}/ w,\\
\sigma_x^2 & = & \sum_{i=-d}^d \sum_{j=-d}^d w_{ij}(x_{ij}-\mu_x)^2/ w,\\
\sigma_{xy} & = & \sum_{i=-d}^d \sum_{j=-d}^d w_{ij}(x_{ij}-\mu_x)(y_{ij}-\mu_y)/ w,
\end{eqnarray}
where $x_{ij}$ is the pixel value at $(i,j)$ in $x$. $w_{ij}=e^{\frac{-(i^2+j^2)}{2\rho^2}}$ and $w=\sum_{i=-d}^{i=d}\sum_{j=-d}^{j=d}w_{ij}$.  $d$ represents the radius of the patch. We use $\rho=1.5$ and $d=5$ here. The SSIM between two images can be obtained through averaging the values of (\ref{eq:ssim}) at all positions. For any image sequence, the average SSIM across all frames is denoted as SSIM$_{\mathrm{vh}}$. To measure the quality of recovered temporal structures, we also slice a video along the horizontal axis and compute average SSIM of all images spanned by the vertical and temporal axes, denoted as SSIM$_{\mathrm{vt}}$.

\begin{figure}[t]
\centering
\begin{tabular}{p{0.205\linewidth}<{\centering}p{0.205\linewidth}<{\centering}
p{0.205\linewidth}<{\centering}p{0.205\linewidth}<{\centering}}
 \scriptsize SRResACNN & \scriptsize ESRCNN  & \scriptsize Ours & \scriptsize GT \\
\end{tabular}
\begin{minipage}{\linewidth}
\centering
\includegraphics[width=0.24\textwidth]{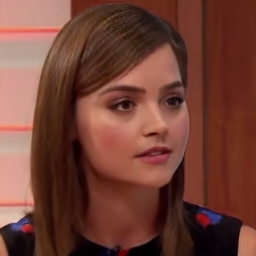}
\includegraphics[width=0.24\textwidth]{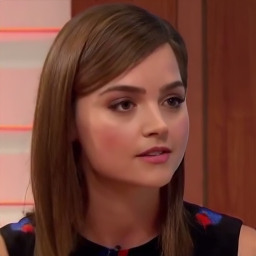}
\includegraphics[width=0.24\textwidth]{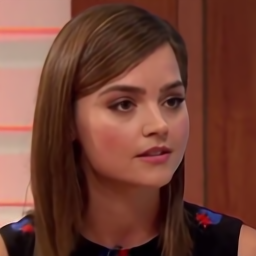}
\includegraphics[width=0.24\textwidth]{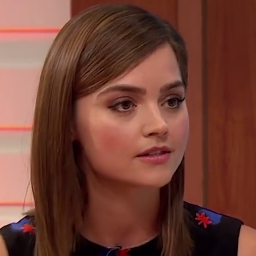}
\end{minipage}\\ 
\begin{tabular}{p{0.2\linewidth}<{\centering}p{0.2\linewidth}<{\centering}
p{0.2\linewidth}<{\centering}p{0.2\linewidth}<{\centering}}
\scriptsize 34.33/0.9151 & \scriptsize 34.40/0.9138 & \scriptsize 35.95/0.9408 & \scriptsize PSNR/SSIM \\
\end{tabular} \\ \vspace{1mm}

\begin{minipage}{\linewidth}
\centering
\includegraphics[width=0.24\textwidth]{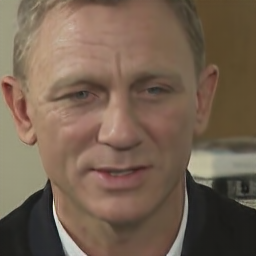}
\includegraphics[width=0.24\textwidth]{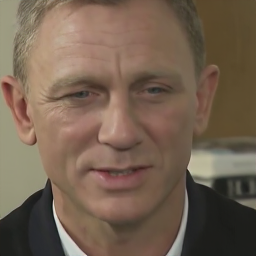}
\includegraphics[width=0.24\textwidth]{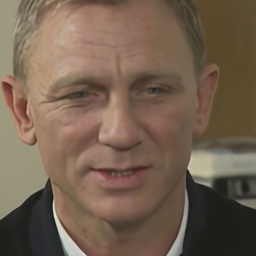}
\includegraphics[width=0.24\textwidth]{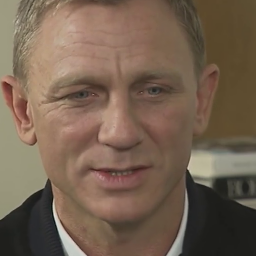}
\end{minipage}\\ 
\begin{tabular}{p{0.2\linewidth}<{\centering}p{0.2\linewidth}<{\centering}
p{0.2\linewidth}<{\centering}p{0.2\linewidth}<{\centering}}
\scriptsize 35.53/0.9414 & \scriptsize 35.96/0.9422 & \scriptsize 39.40/0.9726 & \scriptsize PSNR/SSIM \\
\end{tabular}

\caption{Comparison with GAN-based methods. Both SRResACNN~\cite{pathak2018efficient} and ESRCNN~\cite{wang2018esrgan} are trained under the guidance of GAN as introduced in their original papers. (Best viewed in close-up)}
\label{fig:compgan}
\end{figure} 

\begin{figure*}[t]
\centering
\includegraphics[width=\textwidth]{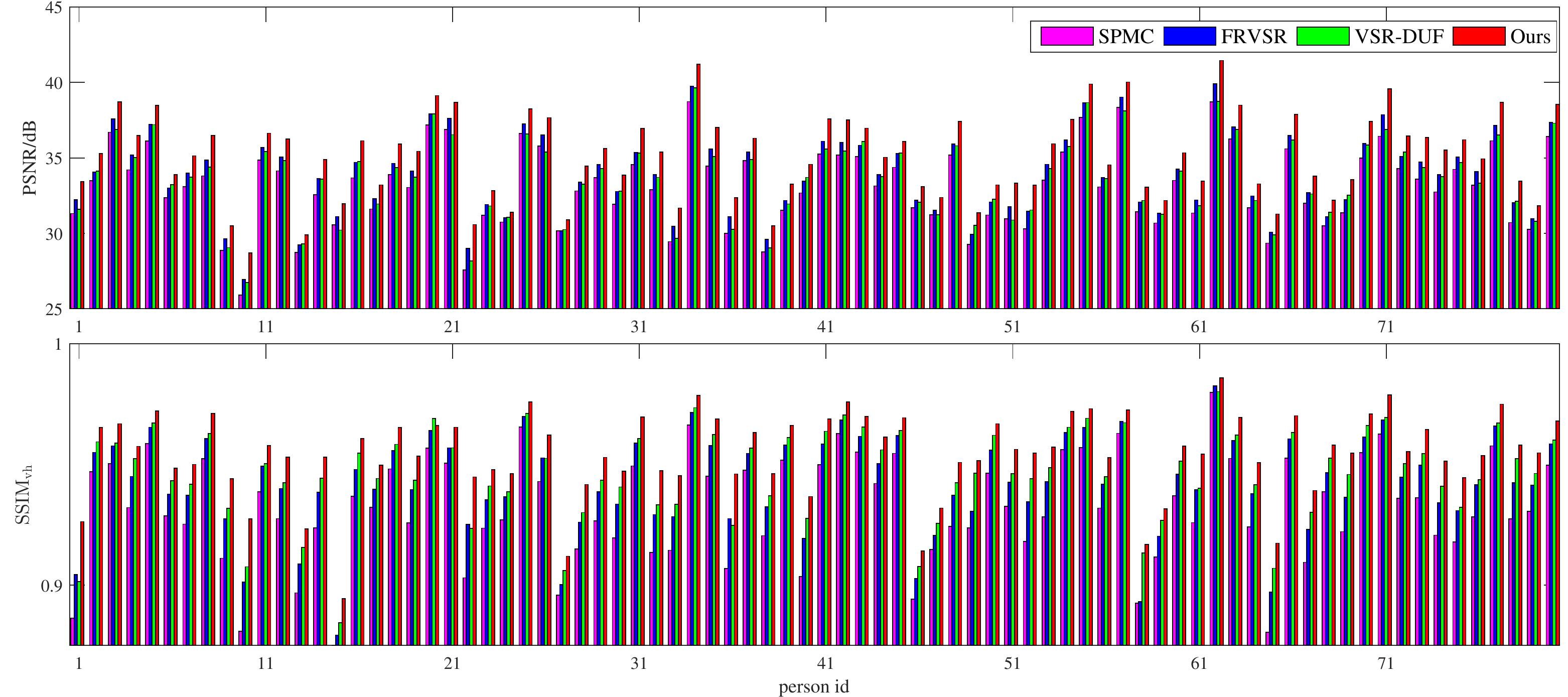}
\caption{Comparison of averaged PSNR and SSIM in facial videos from individual persons. All 80 persons in the testing set of VoxCeleb are considered. Our method performs consistently better than SPMC, FRVSR and VSR-DUF.} \label{fig:avgperson}
\end{figure*}
\subsection{Quantitative and Qualitative Analysis}
\begin{figure*}[t]
\centering
\includegraphics[width=\textwidth]{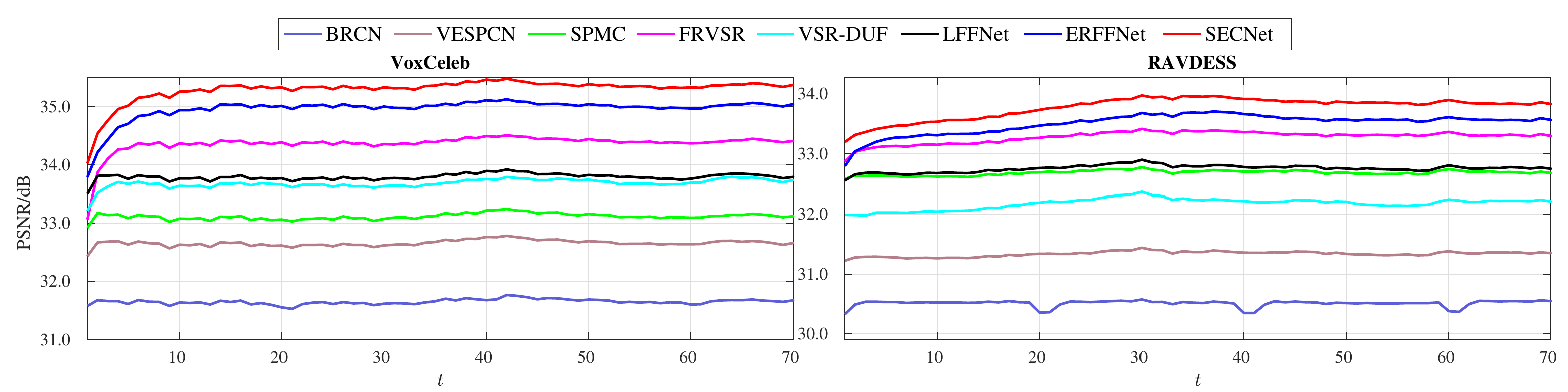}
\caption{Comparison of averaged PSNR at different frames. The PSNR of our method rises quickly as it runs forward from the beginning. Afterwards it steadily maintains at a higher PSNR  rate than other methods.} \label{fig:avgframe}
\end{figure*}

\begin{table*}[t]
\caption{Performance of generic video super-resolution on the testing sets of Harmonic and VID4.}\label{tab:vsr}
\centering
\fontsize{8pt}{10pt}
\selectfont
\setlength\tabcolsep{10pt}
\begin{tabular}{ l||l|l|l||l|l|l }\specialrule{.1em}{0em}{0em}
\multicolumn{1}{c||}{\multirow{2}{*}{}} &  \multicolumn{3}{c||}{Harmonic} & \multicolumn{3}{c}{VID4} \\ \cline{2-7}
& PSNR ($F_t$/$F_p$) & SSIM$_{\mathrm{vh}}$ ($F_t$/$F_p$) & SSIM$_{\mathrm{vt}}$ & PSNR ($F_t$/$F_p$) & SSIM$_{\mathrm{vh}}$ ($F_t$/$F_p$) & SSIM$_{\mathrm{vt}}$ \\ \hline
Bicubic
&29.02(71.5/0.00) &0.8075(70.4/0.00) &0.8245 &22.18(12.4/0.00) &0.6125(21.9/0.00) &0.6860  \\
BRCN~\cite{huang2015bidirectional}
&30.45(46.9/0.00) &0.8415(49.6/0.00) &0.8511 &22.86(9.37/0.00) &0.6730(17.5/0.00) &0.7337 \\
VSRNet$^*$~\cite{kappeler2016video}
&- &- &- &23.26(6.96/0.00) &0.6809(16.5/0.00) &0.7446 \\
VESPCN$^\ast$~\cite{caballero2017real}
&- &- &- &23.66(5.20/0.00) &0.7038(14.5/0.00) &0.7612 \\
VESPCN~\cite{caballero2017real}
&31.04(36.9/0.00) &0.8553(39.3/0.00) &0.8627 &23.44(6.77/0.00) &0.7143(14.3/0.00) &0.7668 \\
SPMC$^\ast$~\cite{tao2017detail}
&- &- &- &24.51(1.78/0.08) &0.7583(8.94/0.00) &0.8025 \\
SPMC~\cite{tao2017detail}
&32.13(19.9/0.00) &0.8787(20.9/0.00) &0.8821 &24.18(3.57/0.00) &0.7670(7.93/0.00) &0.8069 \\

VSR-LTD$^*$~\cite{Liu_2017_ICCV}
&- &- &- &24.01(3.83/0.00) &0.7323(11.6/0.00) &0.7865 \\
FRVSR~\cite{sajjadi2018frame}
&32.24(18.0/0.00) &0.8818(18.1/0.00) &0.8858 &24.21(3.49/0.00) &0.7742(6.74/0.00) &0.8141  \\
VSR-DUF$^\ast$~\cite{jo2018deep}
&32.52(13.7/0.00) &0.8901(11.0/0.00) &0.8916 &24.23(4.00/0.00) &0.8017(3.04/0.00) &0.8328  \\ \hline
LFFNet
&31.82(24.7/0.00) &0.8724(26.1/0.00) &0.8766 &23.96(4.64/0.00) &0.7552(9.79/0.00) &0.7987  \\
SECNet
&\textbf{33.42(0.00/1.00)} &\textbf{0.9026(0.00/1.00)} &\textbf{0.9045} &\textbf{25.02(0.00/1.00)} &\textbf{0.8179(0.00/1.00)} &\textbf{0.8487}  \\
\specialrule{.1em}{0em}{0em}
\end{tabular}
\end{table*}
\subsubsection{Comparisons against State-of-the-Art Methods} Comparisons between our final model SECNet and other state-of-the-art methods  are presented in Table~\ref{tab:facesr} and~\ref{tab:vsr}. In the facial video hallucination task (Table~\ref{tab:facesr}), we compare our proposed method with several state-of-the-art SR methods including GLN~\cite{tuzel2016global}, LapSRNet~\cite{lai2017deep}, \cite{bulat2018super}, SRResCNN~\cite{ledig2017photo}, SRResACNN~\cite{pathak2018efficient}, ESRCNN~\cite{wang2018esrgan}, BRCN~\cite{huang2015bidirectional}, VESPCN~\cite{caballero2017real}, SPMC~\cite{tao2017detail}, FRVSR~\cite{sajjadi2018frame} and VSR-DUF~\cite{jo2018deep}. All methods are trained using the same datasets and settings as described in Section~\ref{sec:data}, except for these marked with `$\ast$' which adopt results released by the authors or generated by provided models. To avoid defects nearby the image borders, input LR images of VSR-DUF are padded with 2 pixels. We conduct  T-test between every contrast method and our proposed method (the last row of the table), to indicate improvement significance. The t-statistic $F_t$ and p-value $F_p$ are presented in the parentheses after PSNR and SSIM$_\mathrm{vh}$. Our SECNet surpasses all previous methods. Practically it outperforms the second best method FRVSR by 2.7\% higher PSNR and 1.0\% larger SSIM$_\mathrm{vh}$ on VoxCeleb.

Comparison of our method against other video super-resolution methods in Harmonic and VID4 datasets is reported in Table~\ref{tab:vsr}. The most peripheral 8 pixels are excluded when computing PSNR and SSIM-s. The self-learned attention is not used in this task. Again our method achieves the best performance. The PNSR and SSIM$_{\mathrm{vh}}$ of our model SECNet are respectively 0.90 and 0.0125 larger than those of the second best method VSR-DUF in Harmonic dataset.

To discuss the efficacy brought by LFFNet, we also transform the enhanced recurrent frame fusion module into an independent SR model called ERFFNet. Apparently it is inferior to SECNet as reported in Table~\ref{tab:facesr}.

A qualitative comparison of facial video hallucination between our method and other SR methods are shown in Fig.~\ref{fig:comp}. The super-resolved results from our method tend to be more appealing and clearer than those from other methods especially in the eye regions. The super-resolving quality in generic single-shot video datasets is shown in Fig.~\ref{fig:compvsr}. Our method recovers the buildings (top image), digits (middle image) and tiles (bottom image) more accurately. Comparison with GAN-based methods \cite{pathak2018efficient} and \cite{wang2018esrgan} is presented in Fig. \ref{fig:compgan}.

\subsubsection{Performance across Persons}
 Performance comparison in facial videos from independent persons is presented in Fig. \ref{fig:avgperson}. It indicates our method consistently performs better than SPMC, FRVSR and VSR-DUF across characters.

\subsubsection{Performance across Frames} We also report averaged PSNR for $t\in [1, 70]$ in Fig.~\ref{fig:avgframe}, where the significance and efficacy of our self-enhanced convolutional network can be clearly observed. 
The PSNR-s of FRVSR and our proposed models rise rapidly during the initial frames because both reuses the estimated results of preceding frames recurrently. However the rising period of SECNet is longer than FRVSR. Overall, our method achieves the highest performance among all considered state-of-the-art SR methods.

\begin{table}[t]
\caption{Comparisons of different fusion strategies on VoxCeleb.}\label{tab:ablation}
\centering
\fontsize{8pt}{10pt}
\selectfont
\begin{tabular}{ c|r||l|l }\specialrule{.1em}{0em}{0em}
\emph{rff} & \emph{sfe} & PSNR($F_t$) & SSIM$_{\mathrm{vh}}$($F_t$) \\ \hline 
 $\times$ & $\times$ 
 &34.16(32.5) &0.9437(47.6) \\ 
$\checkmark$ & $\times$
&34.78(12.2) &0.9514(15.2) \\
$\times$ & one-way CLSTM
&34.42(24.0) &0.9470(34.3) \\ 
$\times$ & cascaded BCLSTM
&34.55(20.0) &0.9485(27.6) \\ \hline 
$\checkmark$ & non-local attention $T_3=2$
&34.95(6.94) &0.9527(9.47) \\
$\checkmark$ & flow-guided attention $T_3=2$
&34.97(6.57) &0.9527(9.49) \\
$\checkmark$ & fused BCLSTM $T_3=2$
&35.14(1.22) &0.9542(2.54) \\ \hline 
$\checkmark$ & recurrent unit in~\cite{huang2015bidirectional}
&34.91(8.39) &0.9523(11.2) \\
$\checkmark$ & one-way CLSTM
&34.98(6.26) &0.9531(7.33) \\
$\checkmark$ & cascaded BCLSTM $T_3=6$
&35.08(3.06) &0.9536(5.36) \\
$\checkmark$ & fused BCLSTM $T_3=6$ 
&\textbf{35.17(0.00)} &\textbf{0.9547(0.00)} \\
\specialrule{.1em}{0em}{0em}
\end{tabular}
\end{table}

\subsubsection{Discussions of Temporal Fusion Strategies}
The facial video super-resolution performances of our final models using one-way, cascaded and fused BCLSTM-s are presented in Table~\ref{tab:ablation}.  Self-attention is not used in all of our models in this subsection. To study the effectiveness of recurrent frame fusion (abbr. \emph{rff}) and sequential feature encoding (abbr. \emph{sfe}), we trained models without using recurrent frame fusion which means that $T_2$ is set to 0, or not using past features to enhance the feature representation of current frame. 
Compared to the model not using \emph{rff} or \emph{sfe}, adopting any of \emph{rff} and \emph{sfe} brings significant improvement. For example  the adoption of \emph{sfe} (equipped with cascaded BCLSTM) and \emph{rff} gives rise to results with 0.39dB and 0.62dB higher PSNR-s respectively than the version in which neither is utilized. Turning off \emph{rff} or \emph{sfe} causes dramatically drop to all metrics. For example abandoning \emph{rff} leads to decrease of 0.53dB for PSNR, in the framework using cascaded BCLSTM as \emph{sfe} strategy. In conclusion, any of recurrent frame fusion and sequential feature encoding can benefit facial video hallucination independently. Adopting both of them leads to better results as they are able to complement each other. Besides, the bidirectional \emph{sfe} strategies outperform one-way strategies. Qualitative comparison of our method using different temporal fusion strategies is presented in Fig.~\ref{fig:compinner}.

Two alternative temporal fusion strategies are tried to replace the BCLSTM based recurrence module. The convolutional non-local operation \cite{wang2018non} can be applied to exploit temporal dependencies. The feature aggregation method in \cite{zhu17fgfa} can also be applied to fuse temporal features based on attention maps which are calculated between the feature of the reference frame and features of previous frames. Optical flow fields are used to align features of previous frames to the reference frame. The comparison is enclosed in Table~\ref{tab:ablation}. Considering the computation load of the non-local operation, $T_3$ is set to 2. The fused BCLSTM performs better than the above two temporal fusion methods. Additionally, we can replace the CLSTM with the recurrent unit in~\cite{huang2015bidirectional}, forming a variant of our method which produces results with 0.26dB lower PSNR.

\begin{table}[t]
\caption{Comparisons of different attention strategies on VoxCeleb.}\label{tab:cmp-att}
\centering
\fontsize{8pt}{10pt}
\selectfont
\begin{tabular}{ l|l|l }\specialrule{.1em}{0em}{0em}
attention strategy & PSNR($F_t$/$F_p$) & SSIM$_{\textrm{vh}}$($F_t$/$F_p$) \\ \hline
without
&35.17(1.69/0.09) &0.9547(1.68/0.09)   \\
non-local operation
&35.15(2.56/0.01) &0.9543(3.72/0.00)  \\
spatial SE
&\textbf{35.26(-1.16/0.24)} &0.9550(0.54/0.59)  \\
channel-wise \& spatial SE
&35.22(0.00/1.00) &\textbf{0.9551(0.00/1.00)}  \\
\specialrule{.1em}{0em}{0em}
\end{tabular}
\end{table}
\subsubsection{Discussions of Attention Strategies}
We discuss the performance of using pairwise attention calculated with the non-local operation \cite{wang2018non}, channel-wise and spatial attentions computed by squeeze-and-excitation (SE)~\cite{hu2018squeeze}. The non-local operation is integrated into the 4-th and 8-th residual blocks. The quantitative comparisons on VoxCeleb dataset are presented in Table \ref{tab:cmp-att}. Using squeeze-and-excitation based attention achieves better results than using non-local pairwise attention. The spatial attention slightly benefits the super-resolved results while incorporation of additional channel-wise attention fails to bring further improvement.

\begin{table}[t]
\caption{Performances of using different $T_2$ and $T_3$ in the validation set of VoxCeleb.}\label{tab:choice}
\centering
\fontsize{8pt}{10pt}
\selectfont
\setlength\tabcolsep{12pt}
\begin{tabular}{ c|c||c|c|c }
\specialrule{.1em}{0em}{0em}
$T_2$  & $T_3$ &  PSNR & SSIM$_{\mathrm{vh}}$ & SSIM$_{\mathrm{vt}}$\\ \hline
0 & 6 & 35.43 & 0.9596 & 0.9595  \\
1 & 6 & 35.78 & 0.9645 & 0.9641  \\
3 & 6 & 35.93 & 0.9643 & 0.9638  \\ \hline
2 & 0 & 35.82 & 0.9638 & 0.9635  \\
2 & 2 & 36.01 & 0.9647 & 0.9643  \\
2 & 4 & 36.03 & 0.9647 & 0.9644  \\
2 & 6 & \textbf{36.04} & \textbf{0.9649} & \textbf{0.9644}  \\
\specialrule{.1em}{0em}{0em}
\end{tabular}
\end{table}

\subsubsection{Choices for $T_2$ and $T_3$}
The results of choosing different $T_2$ and $T_3$ are reported in Table~\ref{tab:choice}. All experimental results are obtained from testing in the validation set of VoxCeleb. The cascaded BCLSTM is adopted to implement \emph{sfe}. Using 2 previous frames produces better results than using 1 previous frame, but more frames do not help improving super-resolution performance which might be caused by the increased difficulty in learning the dependency between current and previous frames. Increasing $T_3$ from 0 to 2 brings gain of 0.19dB in PSNR. Adopting different values 2, 4 and 6 for $T_3$ leads to almost equivalent performance.

\begin{table}[t]
\caption{Comparisons of variants using different numbers of residual blocks on VoxCeleb.}\label{tab:cmp-nrb}
\centering
\fontsize{8pt}{10pt}
\selectfont
\setlength\tabcolsep{12pt}
\begin{tabular}{ c|l|l|l }\specialrule{.1em}{0em}{0em}
\#residual blocks & PSNR & SSIM$_{\mathrm{vh}}$ & SSIM$_{\mathrm{vt}}$  \\ \hline
2
&35.06 &0.9538 &0.9562  \\
4
&35.18 &0.9544 &0.9568  \\
8
&\textbf{35.26} &\textbf{0.9550} &\textbf{0.9572}  \\
16
&35.24 &0.9548 &0.9571  \\
\specialrule{.1em}{0em}{0em}
\end{tabular}
\end{table}
\subsubsection{Choices for Number of Residual Blocks} We present the performances of using various numbers of residual blocks in Table \ref{tab:cmp-nrb}. Using 8 residual blocks achieves the best performance, while 16 blocks can not bring better results.

\section{Conclusions}
To solve the facial video hallucination problem, we have proposed a self-enhanced convolutional network, which utilizes recurrent frame fusion and sequential feature encoding based on ConvLSTM to take advantage of both spatial and temporal information from past video frames. Furthermore a local frame fusion network is utilized to involve in information from future frames. Our method achieves state-of-the-art performance in both facial video hallucination and more generic single-shot video SR tasks. In the future, it deserves in-depth research to exploit deliberately devised attentions in facial video hallucination, based on facial priors, motion units, expressions, etc. 





\ifCLASSOPTIONcaptionsoff
  \newpage
\fi

\bibliographystyle{IEEEtran}
\bibliography{secn.bbl}

\end{document}